\documentclass[acmsmall]{acmart}
\usepackage{graphicx}
\usepackage{booktabs}
\usepackage{subcaption}
\AtBeginDocument{%
  \providecommand\BibTeX{{%
    \normalfont B\kern-0.5em{\scshape i\kern-0.25em b}\kern-0.8em\TeX}}}

\usepackage{geometry}
\usepackage{graphicx} 
\usepackage{pifont}
\usepackage{xcolor}
\usepackage{soul} 
\usepackage{listings}

\lstdefinelanguage{Python}{
    keywords={and, as, assert, async, await, break, class, continue, def, del, elif, else, except, False, finally, for, from, global, if, import, in, is, lambda, None, nonlocal, not, or, pass, raise, return, True, try, while, with, yield},
    keywordstyle=\color{blue},  
    comment=[l]{\#},
    morecomment=[s]{"""}{"""},
    commentstyle=\color{green!40!black},  
    stringstyle=\color{red},
    basicstyle=\ttfamily\small,
    showstringspaces=false,
    frame=single,
    rulecolor=\color{orange},
    backgroundcolor=\color{gray!10}, 
}

\lstnewenvironment{mintedbox}
{
    \lstset{language=Python}
}
{}

\begin{document}

\title[DroneVis]{DroneVis: Versatile Computer Vision Library for Drones}

\author{Ahmed Heakl}
\email{ahmed.heakl@ejust.edu.eg}
\orcid{0009-0009-8712-1457}
\affiliation{
  \institution{Egypt-Japan University of Science and Technology}
  \city{Alexandria}
  \country{Egypt}
  \postcode{21934}
}
\author{Fatma Youssef}
\email{fatma.youssef@ejust.edu.eg}
\affiliation{
  \institution{Egypt-Japan University of Science and Technology}
  \city{Alexandria}
  \country{Egypt}
  \postcode{21934}
}
\author{Victor Parque}
\email{parque@aoni.waseda.jp}
\affiliation{%
  \institution{Waseda University}
  \department{Department of Modern Mechanical Engineering}
  \city{Tokyo}
  \country{Japan}
  \postcode{169-8050}
}
\author{Walid Gomaa}
\email{walid.gomaa@ejust.edu.eg}
\affiliation{
  \institution{Egypt-Japan University of Science and Technology and Alexandria University}
  \city{Alexandria}
  \country{Egypt}
  \postcode{21934}
}

\renewcommand{\shortauthors}{Ahmed Heakl, et al.}

\begin{abstract}
    This paper introduces DroneVis, a novel library designed to automate computer vision algorithms on Parrot drones. DroneVis offers a versatile set of features and provides a diverse range of computer vision tasks along with a variety of models to choose from. Implemented in Python, the library adheres to high-quality code standards, facilitating effortless customization and feature expansion according to user requirements. 
    In addition, comprehensive documentation is provided, encompassing usage guidelines and illustrative use cases. Our documentation, code, and examples are available in \url{https://github.com/ahmedheakl/drone-vis}.
\end{abstract}  



 \maketitle

\section{Introduction}
\label{sec: Introduction}
    
    \par 
    Drones, also referred to as Unmanned Aerial Vehicles (UAVs), have emerged as invaluable assets in various fields, including agriculture, environmental monitoring, disaster response, and surveillance~\cite{mohsan2023unmanned,madhavan2018unmanned,barbedo2019review}. Their true potential lies in their capacity to comprehend their surroundings and make intelligent decisions. This potential is realized through the integration of computer vision algorithms into drones, enabling them to execute a diverse range of tasks based on the live camera feed they capture~\cite{arafat2023vision,mohsan2023unmanned,akbari2021applications}.

    \par 
    Previous research has focused on enabling object detection and tracking on drones, particularly in surveillance applications. For instance, person detection and tracking is implemented in~\cite{rohan2019convolutional} using a single-shot detector (SSD) on Parrot drones, achieving a remarkable frame rate of 58 frames per second~\cite{rohan2019convolutional}. Another framework, known as Deep Drone~\cite{han2016deep}, incorporated object detection using Faster R-CNN at 1.6 frames per second and detection using the Kernelized Correlation Filter (KFC) algorithm at 70 frames per second~\cite{henriques2012exploiting,henriques2014high}.

    \par 
    While object detection and tracking are crucial functionalities for drones, they represent only a fraction of the essential tasks that these devices can perform. Tasks such as crowd counting for rescue and security operations, action recognition for surveillance applications, depth estimation to enhance spatial awareness, and face detection and pose estimation for surveillance purposes hold equal significance. Unfortunately, prior work predominantly concentrated on implementing specific models for object detection and tracking, neglecting the need for a diverse range of models and tasks tailored to users' specific requirements~\cite{rohan2019convolutional,han2016deep}.

    \par
\begin{figure}[!h]
      \centering
      \includegraphics[width=0.5\textwidth]{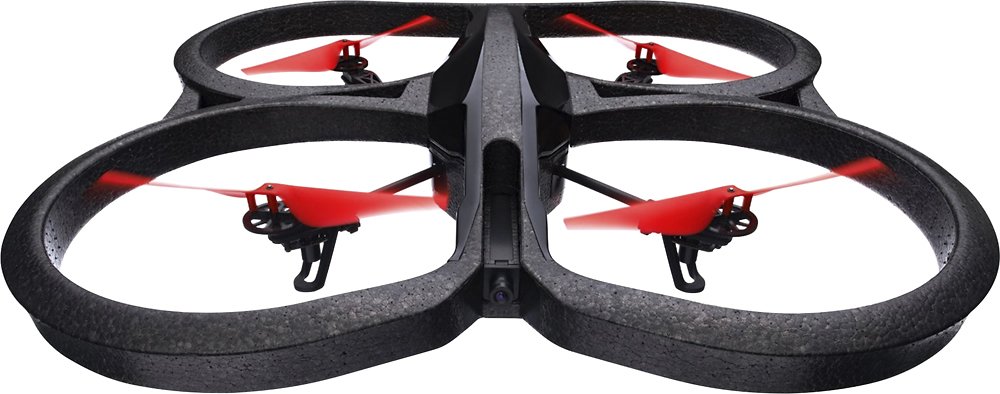}
      \caption{Parrot AR2 drone.}
      \label{fig:Parrot AR2 drone}
    \end{figure}

    Recognizing these gaps, we introduce DroneVis, a versatile library designed to empower researchers and practitioners in the field of drone-based computer vision. DroneVis facilitates the execution of computer vision tasks, especially on Parrot AR2 drones~\cite{parrot2012} depicted in Fig.~\ref{fig:Parrot AR2 drone}. This unmanned aerial vehicle is highly user-friendly, versatile, and cost-effective, making it a prime choice for numerous applications. Featuring an integrated camera with 720p high-definition live video streaming and recording, this drone can capture high-quality visual data for a wide range of applications. Furthermore, its built-in Wi-Fi module allows seamless connectivity to laptops through the drone's Wi-Fi network, offering an impressive communication range of up to 50 meters. 

    
    DroneVis offers a selection of models for each task, providing users with the flexibility to customize their approach according to their preferences. To ensure robustness, DroneVis includes an extensive set of test cases. Additionally, we provide comprehensive documentation that outlines the library's features, installation instructions, and an overview of the various tasks and models it encompasses. 

    \par
    Our main contributions can be summarized in the following points:
    \begin{itemize}
        \item Introducing DroneVis, a versatile library for automating computer vision tasks such as object detection, tracking, segmentation, crowd counting and depth estimation on Parrot drones.
        \item Offering comprehensive documentation for the library and practical usage examples enabling straightforward implementation.
        \item Rigorous testing, ensuring the library's reliability with an impressive 83\% test coverage.
        
        \item User-friendly control options catering to users without extensive programming knowledge.
        \item Development of a gesture recognition model for intuitive remote drone control.
        \item Provision of default models for various computer vision tasks, with the flexibility to customize and change these default models based on user preference.
        \item Development of superior face detection, especially suitable for detecting distant faces in drone applications.
    \end{itemize}

    \par 
    The rest of the paper is organized as follows. Section~\ref{sec: Features} elaborates on the features offered by the DroneVis library. Section~\ref{sec: Drone Control and User Interface} discusses drone control and the available user interfaces. Section~\ref{sec: Models} provides a comprehensive exploration of the array of tasks that DroneVis facilitates, along with the diverse models tailored for each task. In Section~\ref{sec:Comparison to Related Software}, we conduct a comparative analysis with other relevant software solutions. Section~\ref{sec:Vision and Drone Integration} presents an integration example between the drone and computer vision models in autonomous navigation using DroneVis. Finally, Section~\ref{sec:Conclusion and Future Work} serves as the conclusion, shedding light on potential avenues for future research and development.

\section{Library Features}
\label{sec: Features}

    \par 
    In this section, we elaborate on the key features of DroneVis and its contribution to the advancement of computer vision-enabled drone applications.
    
    \subsection{Integrated State-of-the-Art Computer Vision Algorithms}

    \par 
    DroneVis serves as a robust repository of state-of-the-art computer vision models, thereby offering users a versatile toolkit for a wide range of tasks. These models encompass renowned architectures, such as Faster R-CNN, YOLOv8, BlazeFace, just to name a few. With capabilities spanning from object detection/recognition, action recognition and crowd counting to face detection, depth estimation, and pose estimation, the library empowers users to effortlessly select and employ the most suitable model for their specific application requirements.
    
    \subsection{Documentation}
    
        \par 
        DroneVis stands out in documentation with two key features:
        \begin{itemize}
            \item \textbf{Contributor-Friendly Codebase:} Meticulously documented code empowers users to easily customize and extend the library, supporting tasks from hyperparameter tuning to creating new classes.
            
            \item \textbf{Comprehensive User Resources:} Extensive documentation on GitHub\footnote{\url{https://github.com/ahmedheakl/drone-vis}} and ReadTheDocs\footnote{\url{https://drone-vis.readthedocs.io/en/latest/index.html}} includes guides on model adjustments and use-case scenarios, fostering collaboration and contribution within the user community.
        \end{itemize}
    \subsection{Demo for Testing}

        \par 
        DroneVis features a dedicated demo simulation to validate the library's robustness and performance. This application is designed to facilitate rigorous testing and evaluation of the library's real-time data handling, inference accuracy, and detection capabilities. The library achieves a  minimum frames-per-second (fps) rate of 4.5 on an Intel Core 8 CPU, demonstrating its capability to deliver real-time results.    
    \subsection{Test Coverage}
    
        \par 
        DroneVis boasts a high code coverage and employs type hints throughout its implementation to ensure the integrity and reliability of the library's codebase. This quality assurance measure reinforces code robustness and reliability, with code coverage and quality metrics consistently exceeding 80\% with pytest framework~\cite{pytest7.4}.
        
    \subsection{Wide Variety of Drone Control}
            \par 
           DroneVis extends its utility beyond computer vision algorithms by providing comprehensive control over the drone's operations. Users can seamlessly initiate critical actions such as takeoff, landing, reset, emergency response, and calibration. Furthermore, the library equips users with the capability to maneuver the drone along various spatial dimensions, enabling movement in directions encompassing right, left, up, down, forward, and backward.
           
    \subsection{High Quality Implementation}

        \par
        DroneVis adheres to PEP8 coding standards, ensuring a structured and coherent codebase. This commitment enhances readability, simplifies comprehension, and fosters a collaborative development environment, empowering users for confident contributions and library expansion.
        
    \subsection{Multiple User Interfaces}

        \par 
        Recognizing the diverse preferences of users, DroneVis presents a variety of user interfaces to cater to different usage scenarios. Users can seamlessly interact with the library through a graphical user interface (GUI), a command-line interface (CLI), or an innovative gesture-based interface. The latter enables users to control the drone’s actions using intuitive hand gestures, showcasing the library’s commitment to enhancing user experiences. Additionally, for those who prefer more traditional methods, DroneVis also supports control of the drone via keyboard and joysticks.

    \subsection{User Friendly API}

        \par 
        DroneVis provides a user-friendly and intuitive API,  specifically designed to speed up the process of developing customized computer vision applications on drones.
        The API's simplicity empowers users to easily integrate their algorithms with the library's framework.
         As a demonstration, consider the following code snippet, which showcases the ease of utilizing the YOLOv8Detection model for object detection within DroneVis:
        \begin{mintedbox}
        """Example Usage: Object Detection with YOLOv8"""
        from dronevis.models import YOLOv8Detection

        model = YOLOv8Detection() # create mode instance

        model.load_model() # load pre-trained model
        
        model.detect_webcam() # start detection on webcam
        \end{mintedbox}
        
\section {Drone Control and User Interface}
\label{sec: Drone Control and User Interface}

    \par 
    In this section, the available user interfaces and alternative control mechanisms will be discussed in detail, highlighting their features and use cases. Additionally, we explore how the library facilitates connectivity between the laptop and the drone to enable these control mechanisms.
    
    \subsection{Graphical User Interface (GUI)}
    \label{sec: Graphical User Interface (GUI)}
    
        \par
\begin{figure}[!t]
          \centering
          \includegraphics[width=0.8\textwidth]{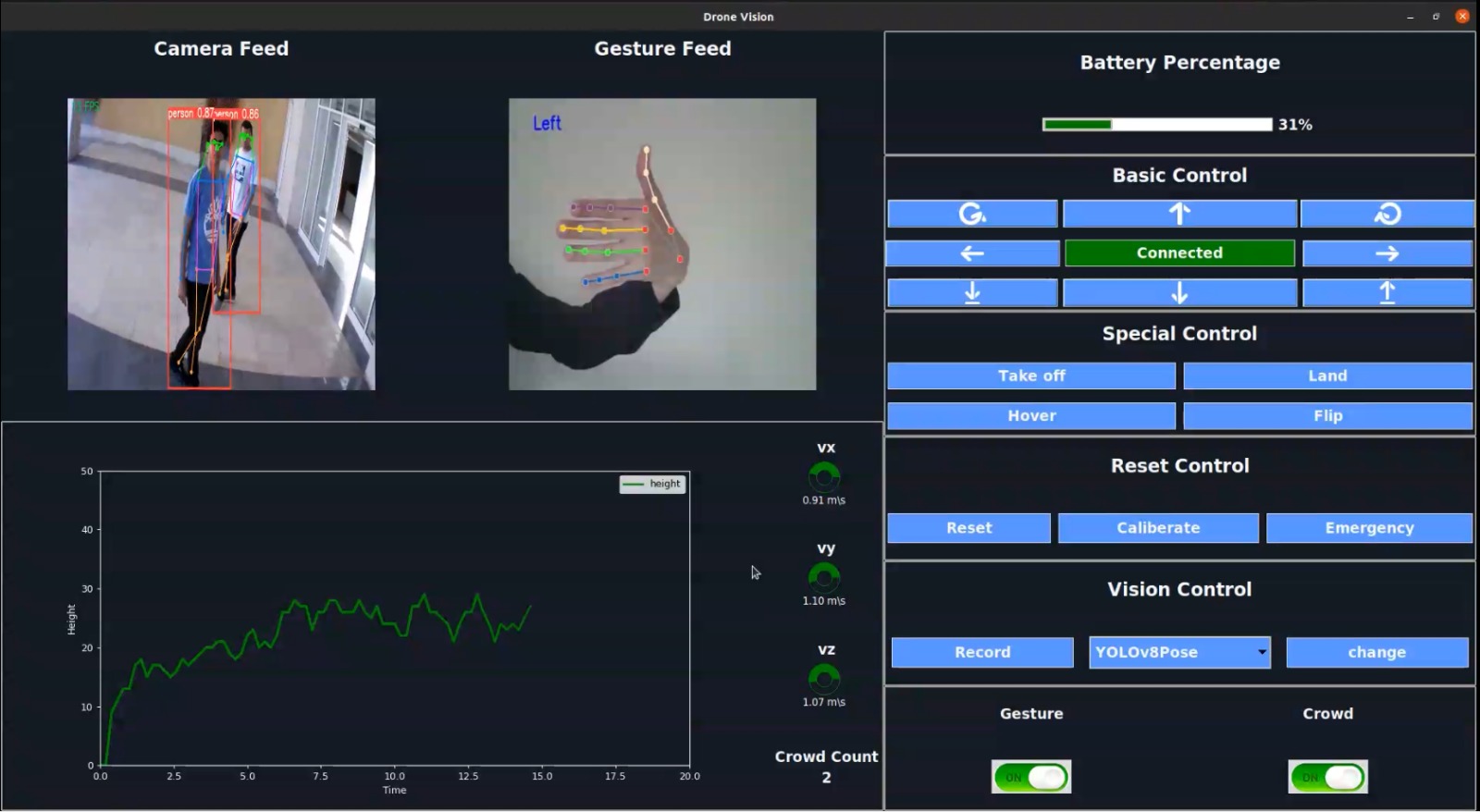}
          \caption{DroneVis GUI.}
          \label{fig:Drone-Vis GUI}
        \end{figure}

        The library is built-in with out-of-the-box GUI to ease the use of the proposed computer vision control and gives the user real-time, full access over their drone. The GUI is built with Tkinter for its ease of use, fast response, and rendering. Moreover, it is a cross-platform library allowing smoother integration with almost all platforms such as Linux, Windows, and MAC.
       To commence, the user should only initiate the following command within the terminal:
       
       \texttt{\$ dronevis-gui}\\
       This action will result in the GUI window opening, as depicted in Fig.~\ref{fig:Drone-Vis GUI}.
       
        The GUI depicted in Fig.~\ref{fig:Drone-Vis GUI} facilitates comprehensive drone management, allowing users to control drone movement through both basic, specialized, and reset control sections. It also provides the capability to execute various computer vision tasks on the drone's camera feed using the vision control section.

    \subsection{Command Line Interface (CLI)}
    \label{sec:Command Line Interface (CLI)}
        \par
\begin{figure}[!t]
          \centering
          \includegraphics[width=0.4\textwidth]{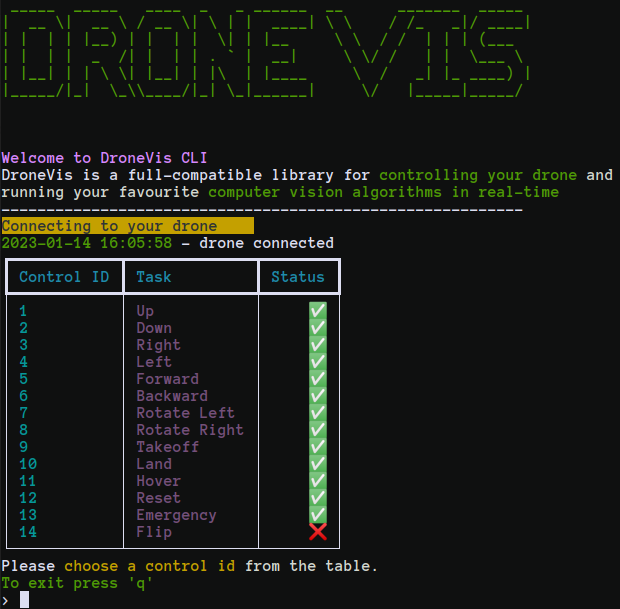}
          \caption{DroneVis CLI.}
          \label{fig:Drone-Vis CLI}
        \end{figure}

        The library also incorporates CLI, a resource-efficient user interface that is favoured for devices with limited computing power. CLI efficiently manages processing and memory resources, ensuring effective communication with the drone. CLI is designed mainly for systems that lack a graphical user interface (GUI) window, making it well-suited for remote services and situations where GUI interactions are not feasible. Unlike GUI, CLI relies solely on text-based commands for controlling drone movement. It doesn't include graphical features, which are common in GUI, especially in the context of vision tasks. To start the CLI, the user should simply write the following command from the terminal:
        
        \texttt{\$ dronevis}\\
        Executing this command will bring up the CLI Interface shown in Fig.~\ref{fig:Drone-Vis CLI}.

    \subsection{Keyboard and Joystick Control}
    \label{sec: Keyboard and Joystick Control }
    \par
        In addition to the aforementioned user interfaces, DroneVis supports drone control via a keyboard and joystick. This control method is suitable for users who are familiar with traditional remote control devices and offers a tactile and responsive way to operate the drone. Keyboard and joystick control is valuable for tasks requiring precise manoeuvring. 
    
    \subsection{Hand Gesture Drone Control}
    \label {sec: Hand Gesture Drone Control}
        \par
\begin{figure*}
        \centering
        \captionsetup[subfigure]{justification=centering}
        \begin{subfigure}{0.3\textwidth}
          \includegraphics[width=\linewidth]{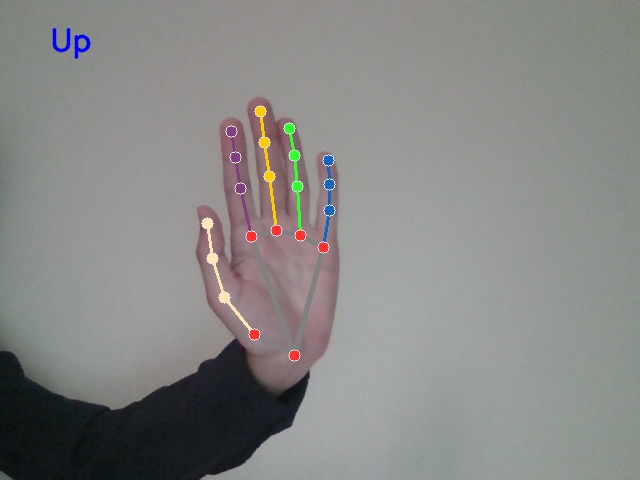}\par
          \caption{}
          \label{fig:sub-first} 
        \end{subfigure}
        \begin{subfigure}{0.3\textwidth}
          \includegraphics[width=\linewidth]{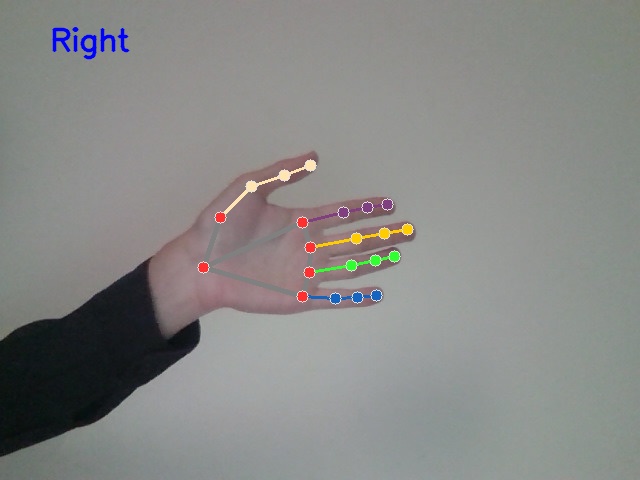} \par 
          \caption{}
          \label{fig:sub-second}
        \end{subfigure}
        \begin{subfigure}{0.3\textwidth}
          \includegraphics[width=\linewidth]{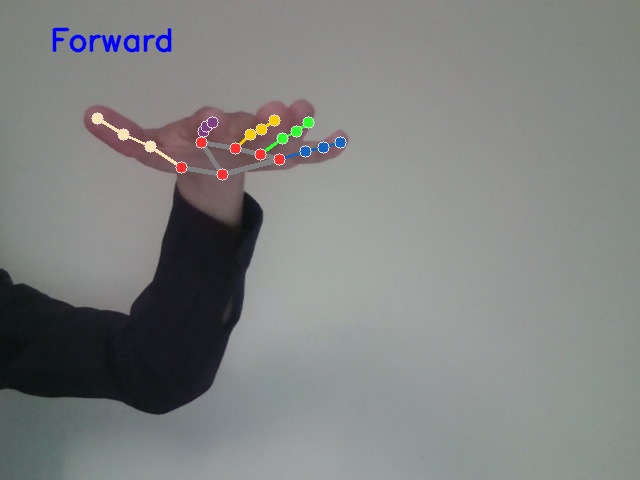}  
          \caption{}
          \label{fig:sub-third}
        \end{subfigure}
        \begin{subfigure}{0.3\textwidth}
          \includegraphics[width=\linewidth]{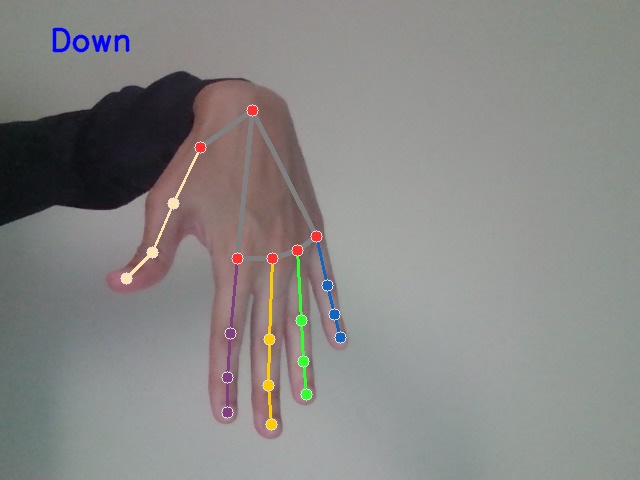}\par 
          \caption{}
          \label{fig:sub-forth}
        \end{subfigure}
        \begin{subfigure}{0.3\textwidth}
          \includegraphics[width=\linewidth]{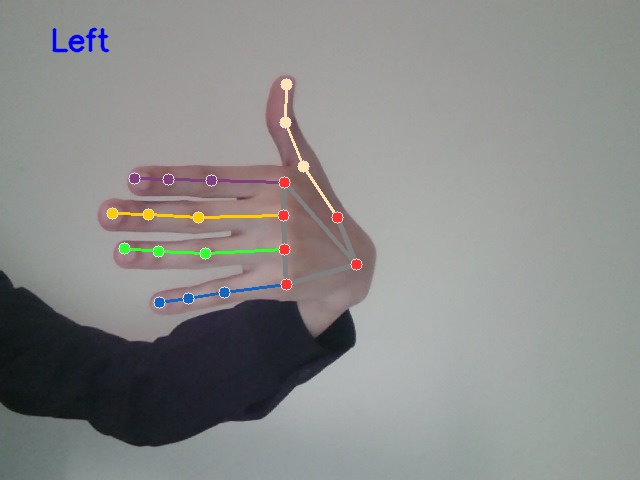}\par  
          \caption{}
          \label{fig:sub-fifth}
        \end{subfigure}
        \begin{subfigure}{0.3\textwidth}
          \includegraphics[width=\linewidth]{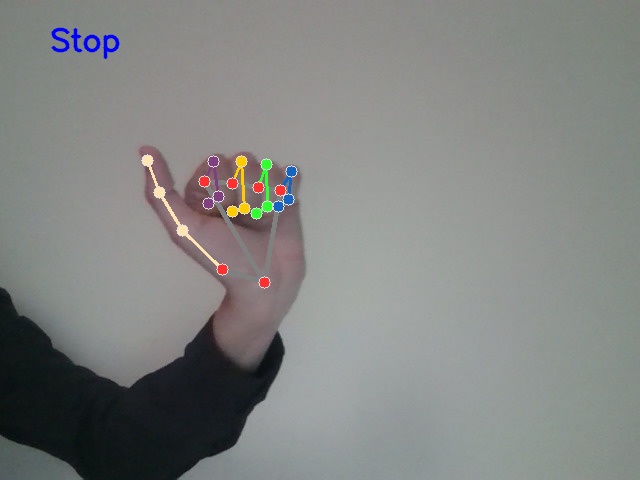}
          \caption{}
          \label{fig:sub-sixth}
        \end{subfigure}
        \caption{Hand gestures for drone control.}
    \label{fig:Hand Gesture for Drone Control}
    \end{figure*}
\begin{figure}[!h]
      \centering
      \includegraphics[width=0.8\textwidth]{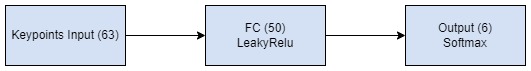}
      \caption{Gesture classifier network.}
      \label{fig:Gesture classifier network}
    \end{figure}

        While user interfaces provide comprehensive access to drone functions, including movement control and various computer vision tasks through the drone's camera, they necessitate proximity to a computer device. In cases where simple drone movement control is sufficient, hand gesture control can offer enhanced intuitiveness and interactivity compared to traditional user interfaces. The DroneVis library enables straightforward drone control using hand gestures, as illustrated in Fig.~\ref{fig:Hand Gesture for Drone Control}. To achieve this capability, the Mediapipe hands model~\cite{zhang2020mediapipe} is employed to extract the keypoints of each hand, yielding 3D coordinates for 21 keypoints. Six distinct gestures are utilized for drone control, as showcased in Fig.~\ref{fig:Hand Gesture for Drone Control}. A small dataset comprising 328 images encompassing these six gestures is collected for training a gesture classifier model, as depicted in Fig.~\ref{fig:Gesture classifier network}. The model's input consists of the 63-dimensional keypoints extracted by Mediapipe ($21 \times 3$), representing the $x$, $y$, and $z$ coordinates of the $21$ keypoints. The classifier network incorporates a single fully connected layer with 50 units, followed by a leaky ReLU activation, and an output layer with 6 units corresponding to the six gestures, followed by a softmax activation. Layer count and sizes are empirically determined for optimal performance, ensuring real-time feasibility for drone control. The dataset is partitioned into 80\% training, 10\% validation, and 10\% testing sets, with label stratification to account for minor class imbalances. An exemplar output classification is demonstrated in Fig.~\ref{fig:Hand Gesture for Drone Control}, showcasing the associated class label.

    \subsection{Drone Connectivity}
    This section focuses on creating a robust connection between the laptop and the drone, facilitating seamless communication during the drone's flight.
    
    Upon linking to the drone's Wi-Fi network, separate sockets are initiated for each of the drone's dedicated ports, each designed for a distinct purpose. Notably, one socket focuses on video streaming, while another handles control commands, and yet another gathers navigation data. These sockets are instrumental for transmitting requests and retrieving data, maintaining a seamless line of communication during the drone's flight. To ensure real-time and responsive control, DroneVis implements separate threads for each connection type—command, video, and navigation data. Each thread aligns with a designated port, enhancing the efficiency of drone communication. Comprehensive details regarding drone connections can be explored in the drone connection section on the ReadtheDocs platform\footnote{\url{https://drone-vis.readthedocs.io/en/latest/connection/drone_connection.html}}.

\section {Tasks and Models}
\label {sec: Models}

    \par
    The DroneVis library offers a comprehensive collection of models designed for diverse computer vision tasks. In this section, we provide an in-depth analysis of each task, elucidating the specific nature and requirements of the task at hand. Furthermore, we present an overview of the neural network models available within the library for each task, where we utilize the available pre-trained weights rather than training the models from scratch, emphasizing their respective strengths and capabilities. Additionally, we evaluate and recommend the most suitable model for each task based on its accuracy and computational efficiency, considering factors such as running time.

    To test the capabilities of the DroneVis library, we conducted a series of experiments utilizing a drone equipped with a high-quality camera of 720p. The drone was flown within the confines of the Egypt-Japan University of Science and Technology campus, capturing real-world footage that served as the basis for numerous computer vision tasks. This real-world data allowed us to evaluate the performance and applicability of the library in practical scenarios.

    For certain tasks, such as object detection, tracking, pose estimation, segmentation, and depth estimation, we relied on our drone footage as the primary source of input data. The high-resolution imagery obtained from the drone's camera was instrumental in these tasks, providing rich visual data for analysis. The data collection process adhered to ethical guidelines and included verbal consent from the individuals involved. Subjects were informed about the research objectives and the intended use of the aerial footage, and their voluntary verbal consent was obtained. The study was conducted in compliance with the research ethics of Egypt-Japan University of Science and Technology. We also respect the terms and conditions of the data usage agreement with the university and the participating individuals.

    However, there were scenarios in which we encountered limitations. Action recognition, for instance, presented challenges as the well-known action recognition dataset include a wide range of complex actions, that could not be feasibly performed within the confines of the university campus. In such cases, we turned to test videos from well-known datasets to assess the library's performance. Additionally, for lane and road detection, we faced regulatory restrictions that prohibited drone flight outside the university campus. Consequently, we could not rely on our drone footage for these tasks, and instead, we utilized test videos from established datasets to measure the models' performance.

    \subsection{Object Detection}
    \label{sec: Object Detection}

        \par 
        Object detection is a fundamental task in computer vision, particularly in the context of Parrot drones. It involves the automatic identification and localization of multiple objects within images or video streams. By employing advanced algorithms and techniques, object detection enables Parrot drones to navigate and avoid obstacles autonomously. The significance of object detection lies in its wide range of applications, such as aerial surveillance, search and rescue operations, agriculture monitoring, and autonomous navigation in complex environments. 
        
        \par
        Several popular object detection models include Faster R-CNN, SSD, and YOLO. Faster R-CNN and SSD are anchor-based models that rely on predefined anchor boxes of varying scales and aspect ratios to identify objects within images. These models perform regression to align the anchor boxes with ground truth bounding boxes, achieving precise localization. In contrast, YOLO (You Only Look Once) operates differently. It divides the image into a grid and predicts bounding boxes and class probabilities for each grid cell.
        
        \par
        In this paper, we delve into the details of the above-mentioned models, and explore their effectiveness in automating computer vision algorithms on Parrot drones. Fig.~\ref{fig: object detection output} illustrates the output of these different object detection models.

         \begin{figure*}
            \centering
            
            \captionsetup[subfigure]{justification=centering}
            \begin{subfigure}{0.45\textwidth}
              \includegraphics[width=\linewidth]{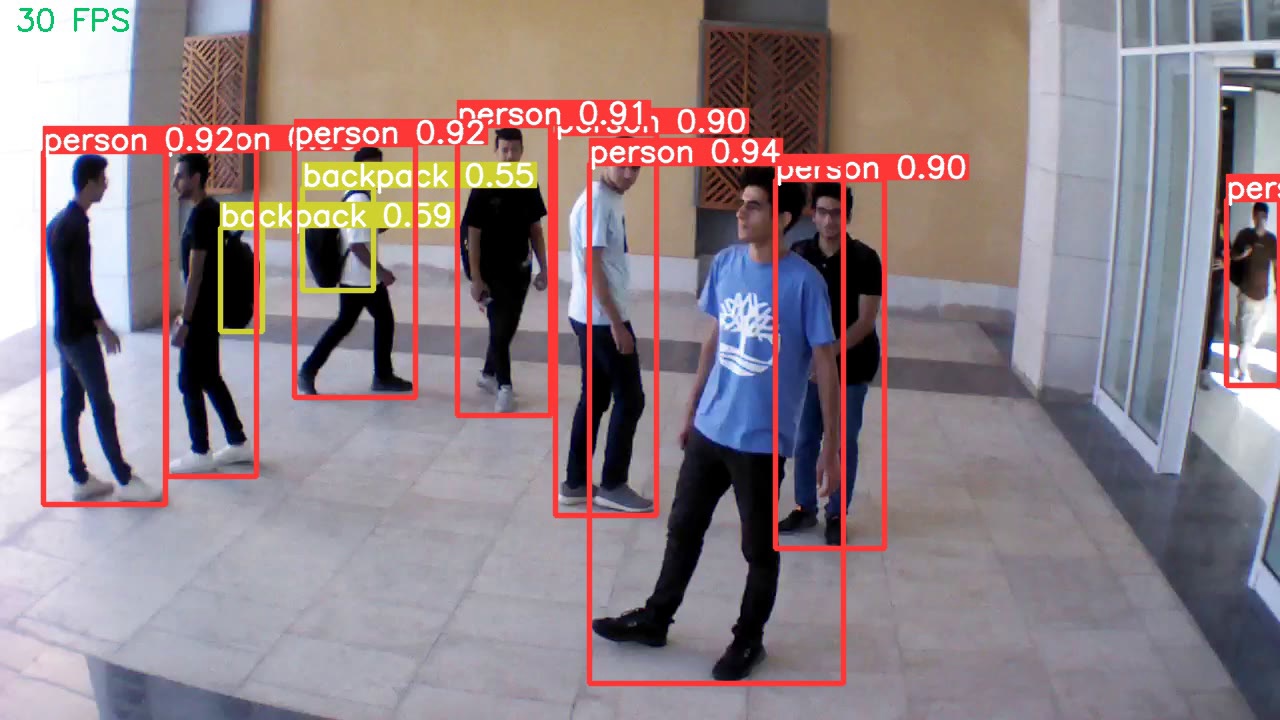}\par
              \caption{YOLOv8}
              \label{fig:sub-first}
            \end{subfigure}
            \begin{subfigure}{0.45\textwidth}
              \includegraphics[width=\linewidth]{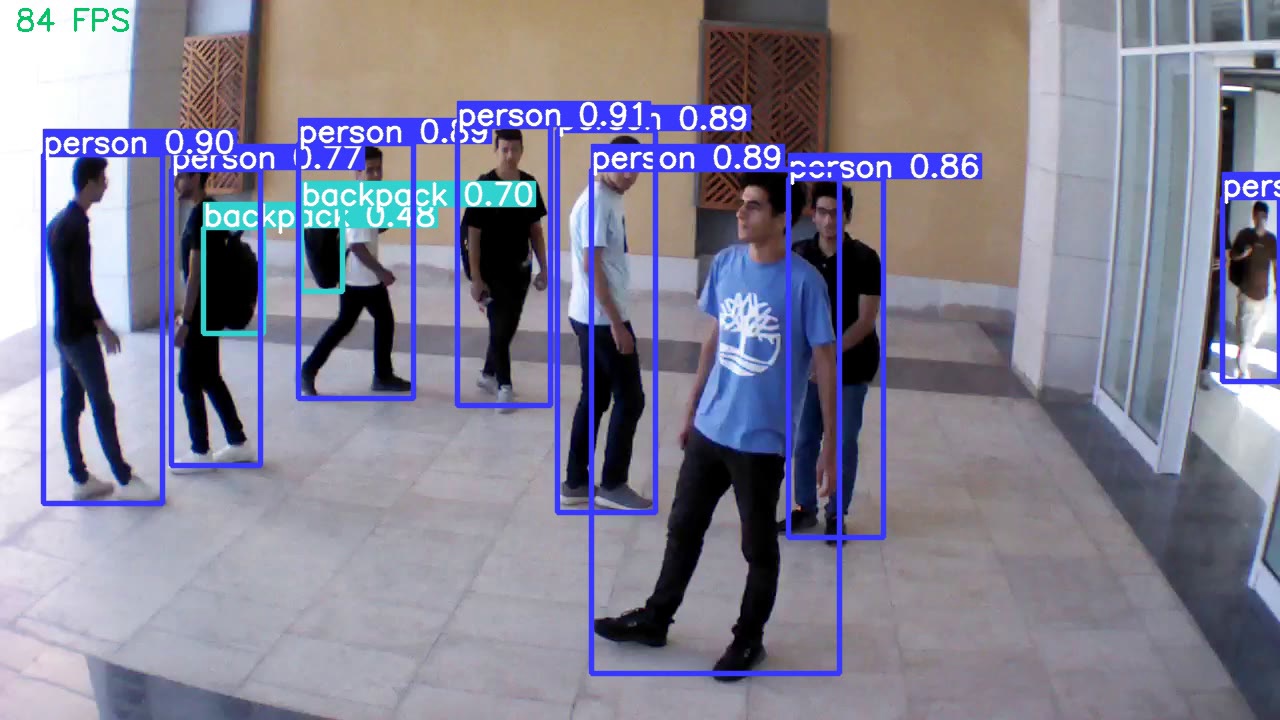} \par 
              \caption{YOLOv5}
              \label{fig:sub-second}
            \end{subfigure}
            \begin{subfigure}{0.45\textwidth}
              \includegraphics[width=\linewidth]{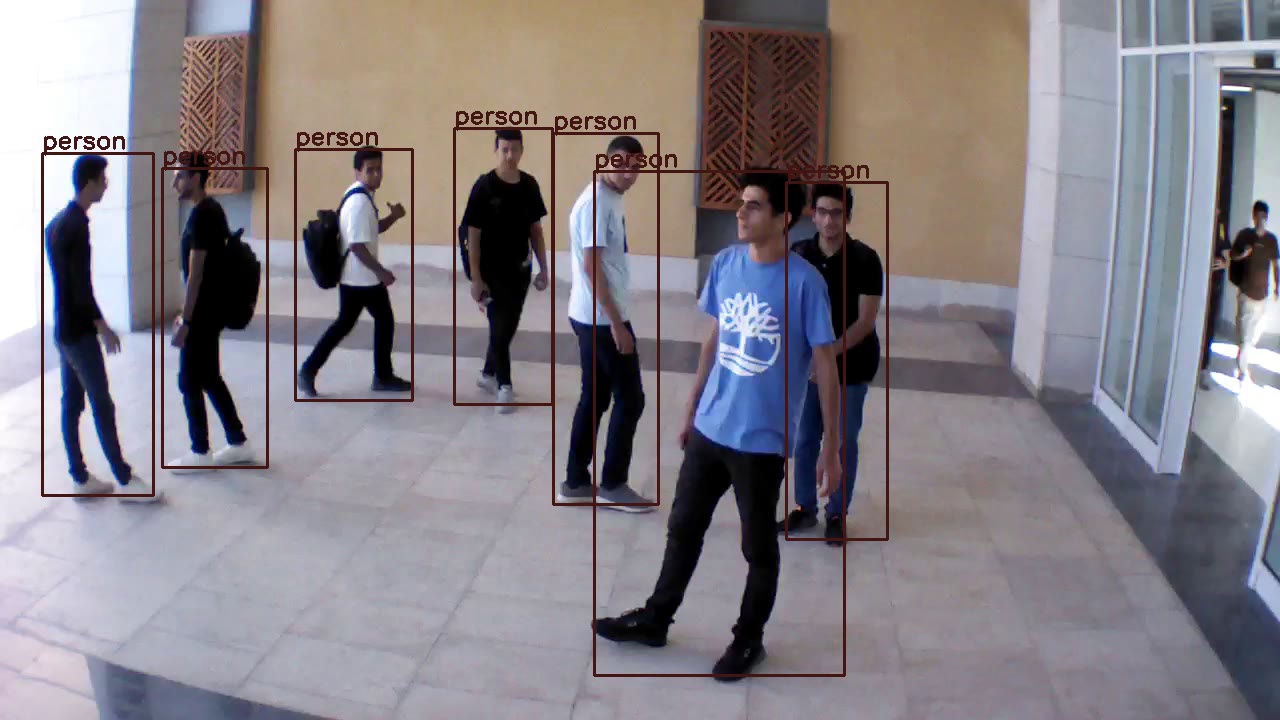}\par 
              \caption{Faster RCNN}
              \label{fig:third-forth}
            \end{subfigure}
            \begin{subfigure}{0.45\textwidth}
              \includegraphics[width=\linewidth]{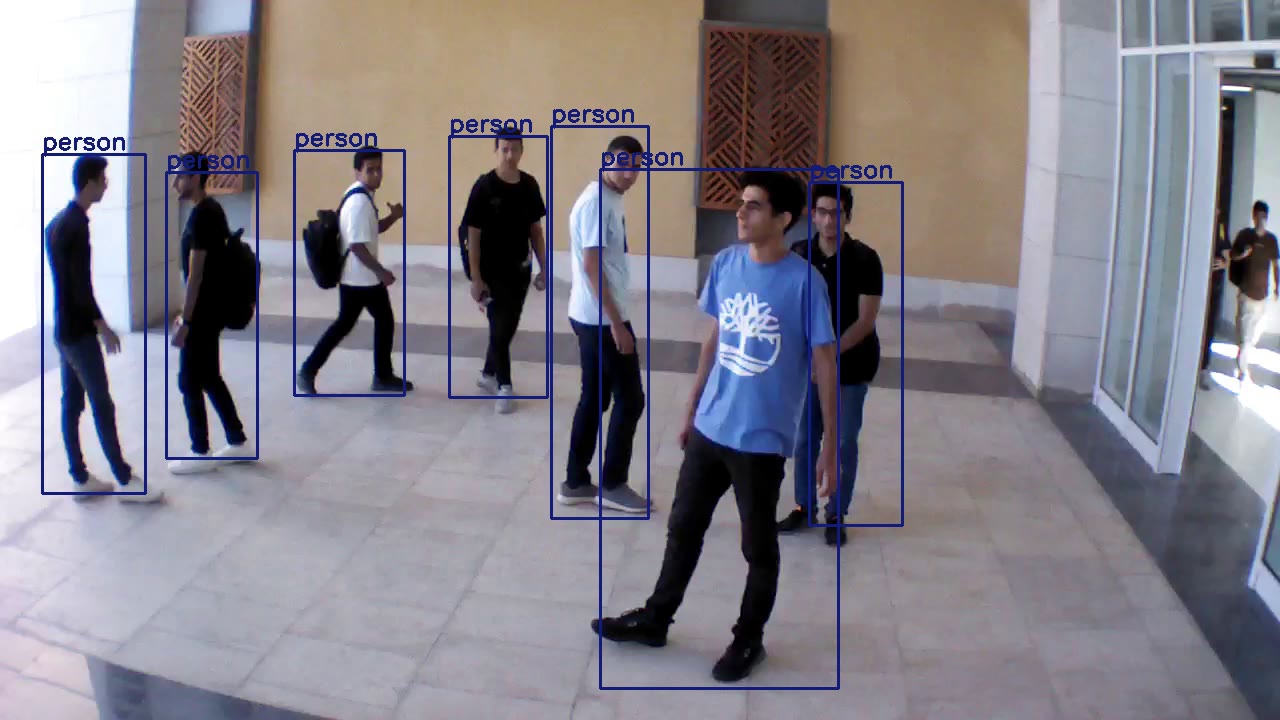}\par  
              \caption{SSD}
              \label{fig:sub-forth}
            \end{subfigure}
    \label{fig: object detection output}
    \caption{\label{fig: object detection output} Output of object detection using various models.}
    \end{figure*}
        
        \subsubsection{Faster R-CNN Model}
        
        \par
        The Faster Regional Proposal CNN (Faster R-CNN)\cite{ren2015faster} improves upon R-CNN\cite{girshick2015fast} and Fast R-CNN~\cite{girshick2014rich} by integrating region proposal generation within the network architecture. Unlike traditional methods using selective search~\cite{uijlings2013selective}, Faster R-CNN introduces a Region Proposal Network (RPN) that shares convolutional layers with the object detection network. The RPN efficiently generates region proposals by sliding a small network window over the feature map, simultaneously predicting objectness scores and refining bounding box coordinates. This architectural enhancement allows end-to-end training and faster inference by sharing convolutional layers between the RPN and the object detection network.

        \subsubsection{Single Shot MultiBox Detector (SSD) Model}
        
        \par
        SSD~\cite{liu2016ssd} innovates by utilizing a single network to predict object bounding boxes and class probabilities directly from feature maps at multiple scales. Through default bounding boxes of various aspect ratios and scales at each feature map location, SSD efficiently detects objects of different sizes. The network integrates initial convolutional layers for low-level feature extraction, auxiliary layers tailored for object detection, and convolutional predictors designed to forecast detections. Default boxes with various aspect ratios contribute to multi-scale feature maps, enhancing the network's object detection capabilities. In contrast to Faster R-CNN, SSD simplifies the pipeline and reduces computational costs by performing object detection in a single pass, eliminating the need for a separate region proposal network.

        \subsubsection{You Only Look Once (YOLO) Model} 
        \par
        YOLO ~\cite{redmon2016you} is the most popular and efficient model in computer vision. Introduced in 2015 to be trained end-to-end, it aimed at real-time object detection and classification. The model family belongs to one-stage object detection models that process an entire image in a single forward pass of a CNN. Unlike two-stage detection models such as R-CNN and its variants – first propose regions of interest and then, classify these regions- YOLO processes the entire image in a single pass, making it exceedingly faster. In this work, we chose to incorporate YOLOv5~\cite{jocher2020yolov5} and YOLOv8~\cite{jocher2023yolov8} for the sake of comparison because of their reliable detection and speed.



        \par
        YOLOv5~\cite{jocher2020yolov5} builds upon the YOLOv3 head network and introduces the EfficientDet backbone network, resulting in significant improvements in detection speed and accuracy. It incorporates dynamic anchor assignment to better fit object sizes, improved data augmentation for challenging conditions, and a modified non-maximum suppression algorithm for more efficient and accurate detection. With its flexible and Pythonic structure, YOLOv5 became the world's state-of-the-art repository for object detection in 2020.
        
        


        \par
        YOLOv8~\cite{jocher2023yolov8} is the last model in the YOLO series (at the time of developing our work), surpassing all of them in both accuracy and speed.
        YOLOv8 introduced minor changes, e.g., removal/addition of some CNN layers and changing the kernel sizes), yet the major change was anchor-free detections. YOLOv8 predicts the center of an object directly instead of the offset from a known anchor box. It is more flexible as it does not require the manual specification of anchor boxes, which can be difficult to choose and can lead to sub-optimal results in previous models of YOLO. In addition, YOLOv8 introduced multiple models for solving other common tasks in computer vision – Instance Segmentation, Image Classification, and Object Tracking.  


        \par 
        Table~\ref{tab: Performance Comparison of Various Models} provides a comparative analysis of the performance of the previously mentioned object detection models, evaluating their speed in frames per second (fps) and accuracy in average precision (AP), which measures how well the model correctly detects and localizes instances. It is evident that, in terms of speed, YOLO v8 outperforms all other models, achieving an impressive frame rate of 80 frames per second (fps), while the slowest model, Faster R-CNN, lags significantly behind at a mere 15 fps. Additionally, YOLO v8 exhibits exceptional detection capabilities, with an average precision of 53.9\%.

    \subsection{Object Tracking}
    \label{sec: Object Tracking}
        \par
        Object tracking is the process of locating a certain object or multiple objects in a video stream overtime while keeping a unique identity to each object. Object tracking has numerous applications such as video surveillance. Multi-object detection composes of detecting the objects in each frame, localizing them, and associating similar objects in different frames. It can be seen that the accuracy of the tracking mainly depends on the detection model used. In this work, we employ YOLOv8~\cite{jocher2023yolov8}, offering a tracking mode through two cutting-edge algorithms: BoT-SORT~\cite{aharon2022bot} and ByteTrack~\cite{zhang2022bytetrack}.

        \par
\begin{figure}[h]
        \centering
        \captionsetup[subfigure]{justification=centering}
        \begin{subfigure}{0.45\textwidth}
          \includegraphics[width=\linewidth]{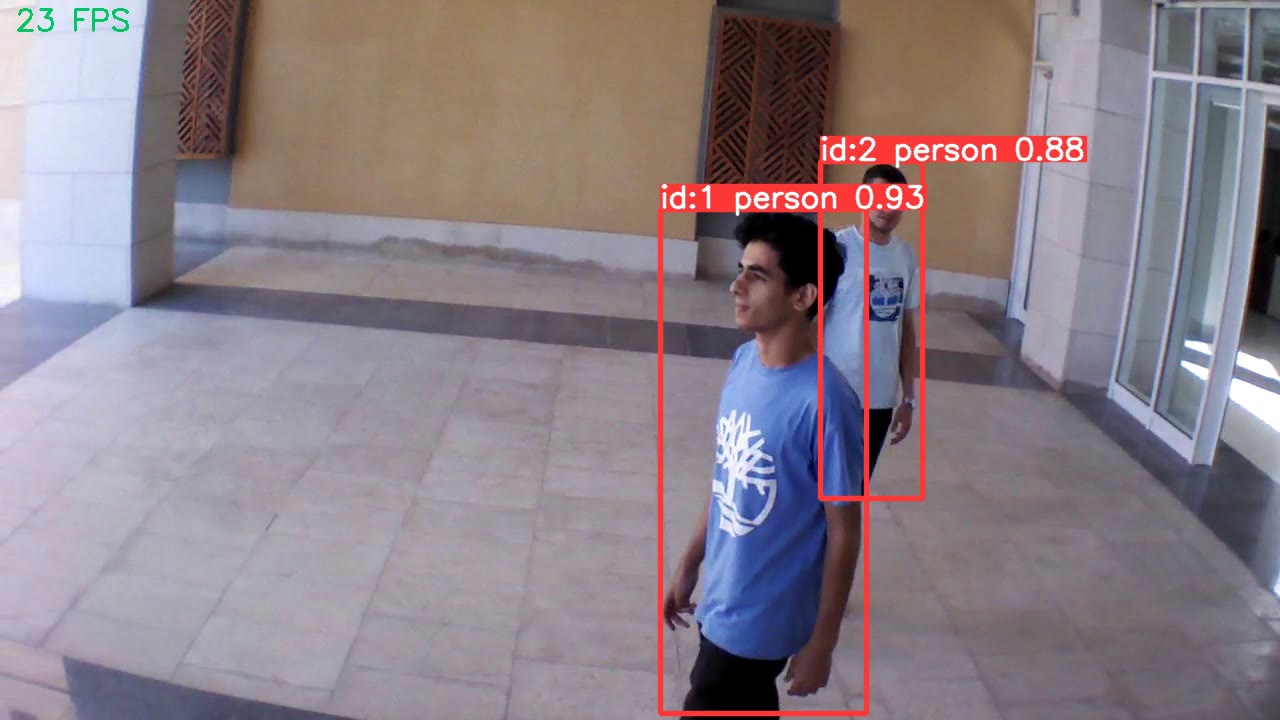}\par
          \caption{First frame.}
          \label{fig:sub-first}
        \end{subfigure}
        \begin{subfigure}{0.45\textwidth}
          \includegraphics[width=\linewidth]{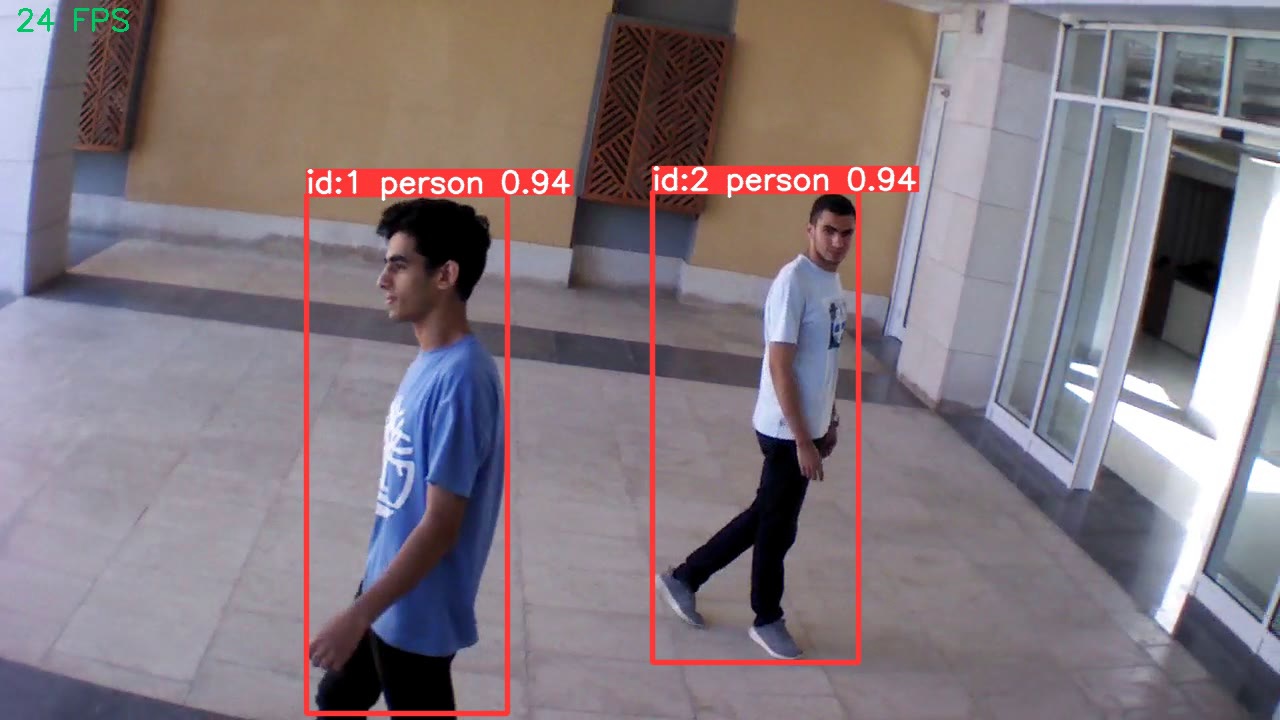} \par 
          \caption{Twentieth frame.}
          \label{fig:sub-second}
        \end{subfigure}
        \caption{\label{fig:Output of tracking using YOLOv8} Output of tracking using YOLOv8. }
        \end{figure}

 
        ByteTrack stands out from other tracking algorithms as it utilizes all detected bounding boxes, even the ones with low scores, enabling recovery of objects with low score detection. It leverages the Kalman filter for predicting new box locations and associates bounding boxes with predicted tracklets, prioritizing high score boxes.
        BoT-SORT builds upon ByteTrack and improves accuracy by employing a more precise Kalman filter and incorporating camera motion compensation, effectively reducing box identity switches. Experiments on the MOT17 benchmark yield a Multi-object Tracking Accuracy (MOTA) of 80.6 for both ByteTrack and BoT-SORT algorithms. Fig.~\ref{fig:Output of tracking using YOLOv8} shows the output of BoT-SORT tracker (default tracker for YOLO v8) in the first and twentieth frames.

    \subsection{Pose Estimation}
    \label{sec: Pose Estimation}
    
        \par
\begin{figure}[htbp]
        \centering
        \captionsetup[subfigure]{justification=centering}
        \begin{subfigure}{0.45\textwidth}
          \includegraphics[width=\linewidth]{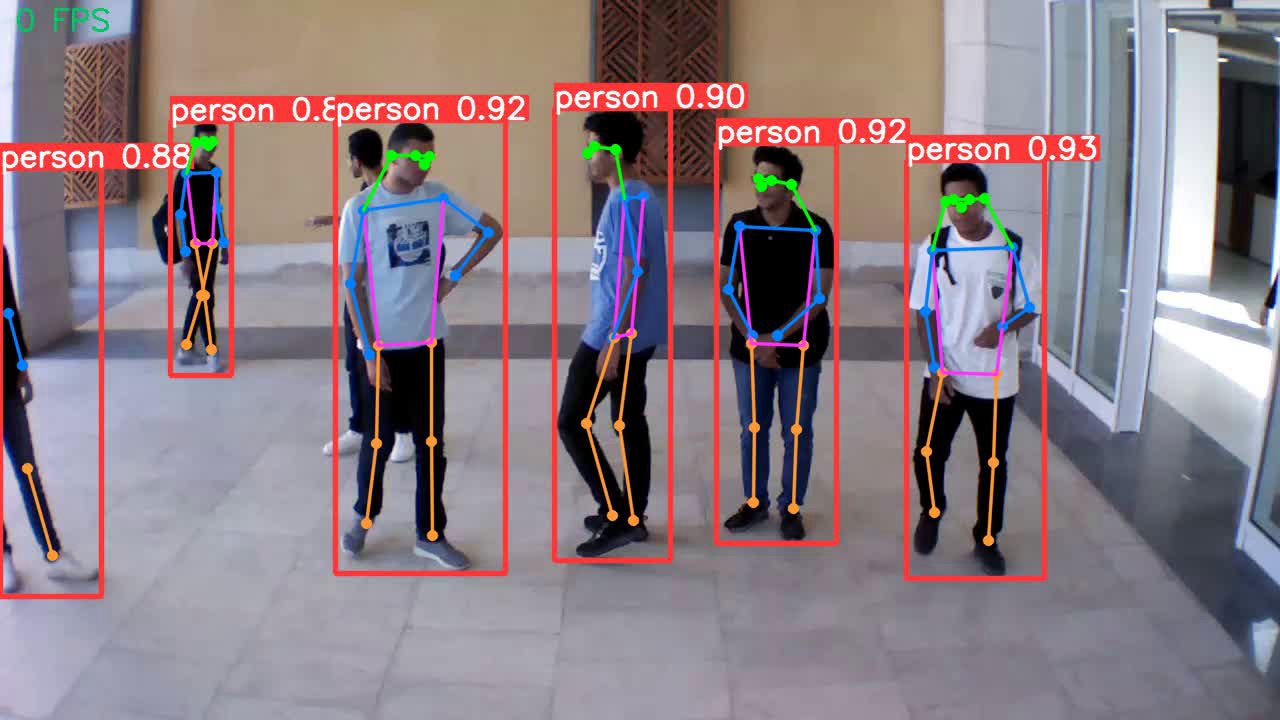}\par
          \caption{Yolov8 Pose.}
          \label{fig:pose_yolov8}
        \end{subfigure}
        \begin{subfigure}{0.45\textwidth}
          \includegraphics[width=\linewidth]{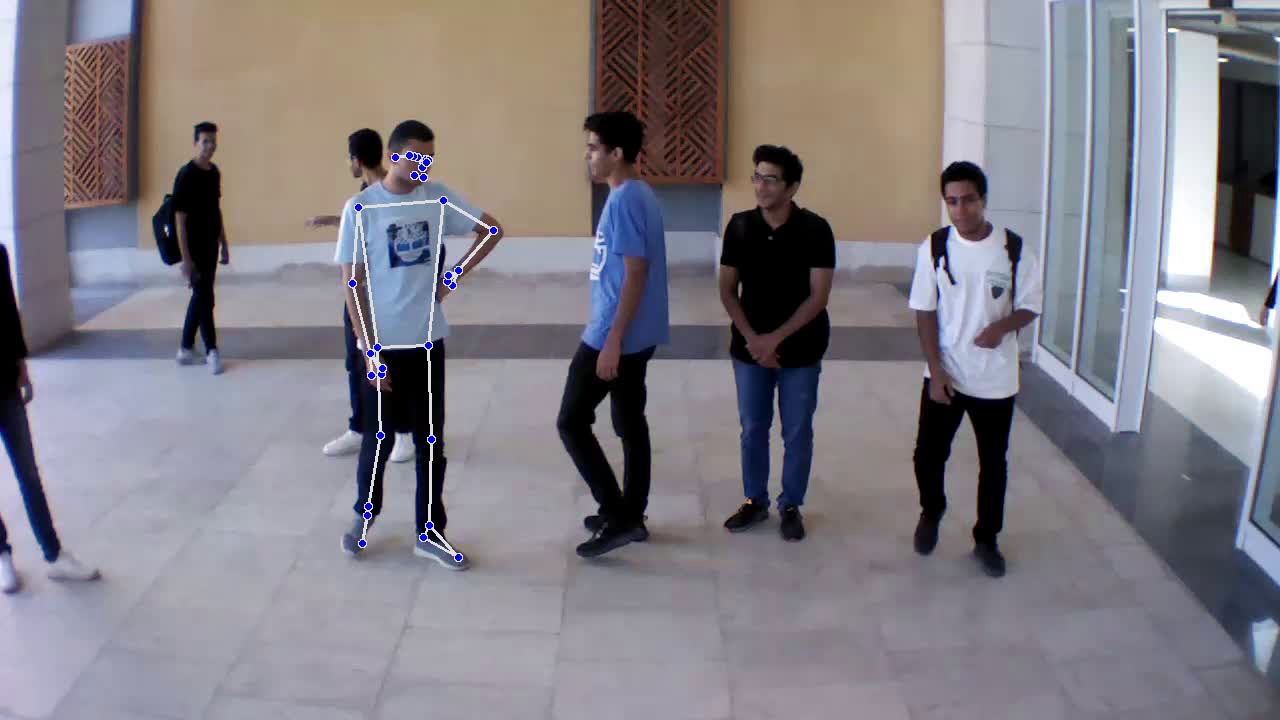} \par 
          \caption{Mediapipe Pose.}
          \label{fig:mediapipe}
        \end{subfigure}
        \caption{\label{fig:Output of pose estimation.} Output of pose estimation.}
        \end{figure}

        Pose estimation aims to detect the positions of specific points, known as keypoints, within an image. These keypoints often correspond to significant features, such as joints, landmarks, or distinctive elements of an object. The positions of these keypoints are typically represented using 2D or 3D coordinates. The outcome of a pose estimation model consists of the keypoints on the object within the image, often accompanied by confidence scores for each point. Pose estimation is an ideal solution when the objective is to recognize particular parts of an object within a scene and understand their spatial relationships with respect to each other. For this task, we offer two state-of-the-art models in pose estimation: YOLOv8-pose~\cite{jocher2023yolov8} and MediaPipe pose estimation~\cite{bazarevsky2020blazepose}. An example of the output of YOLOv8 pose model and MediaPipe model are shown in Fig.~\ref{fig:Output of pose estimation.}.
        
        \subsubsection{YOLOv8-Pose}
        
        \par
        Another feature provided by YOLOv8 is pose estimation. For this task, YOLOv8 was pretrained on the COCO dataset~\cite{lin2014microsoft}. The model outputs 17 2D keypoints with a mAP
        of $50.4$. The mAP metric quantifies how effectively the model's predicted masks align with the ground truth masks for each object instance within the image. The model runs in real-time and accurately estimates the pose even in crowd scenes. 
        
        \subsubsection{MediaPipe Pose Estimation}
        \label{sec:MediaPipe Pose Estimation }
        
        \par
        MediaPipe Pose Estimation is based on the Blazepose architecture~\cite{bazarevsky2020blazepose}. Unlike YOLOv8-Pose, MediaPipe provides 33 3D keypoints in real-time. These keypoints are a superset of the 17 keypoints provided by YOLOv8 (COCO dataset keypoints), and they also include keypoints for the face, hands, and feet (found in BlazeFace~\cite{bazarevsky2019blazeface} and BlazePalm~\cite{zhang2020mediapipe}). The pipeline of this pose estimation involves first detecting a person in the image using a face detector and then predicting the keypoints, assuming that the face is always visible. MediaPipe Pose Estimation is mainly designed for fitness applications for a single person or a few people in the scene so it suffers in detecting multiple persons in an image as shown in Fig.~\ref{fig:mediapipe}.
        
    \subsection{Instance Segmentation}
    \label{sec: Instance Segmentation}
    
        \par
        Instance segmentation goes beyond object detection by not only identifying individual objects in an image but also accurately segmenting them from the surrounding background. The result of an instance segmentation model is a collection of masks or contours outlining each object in the image, accompanied by class labels and confidence scores for each object. Instance segmentation is particularly valuable when detailed information about object boundaries and shapes is necessary, in addition to their spatial locations within the image~\cite{yao2020video}. For this task, DroneVis offers two models: YOLOv8 segmentation and MediaPipe segmentation. The results produced by these two models are visually depicted in Fig.~\ref{fig:Instance segmentation output}.

      \begin{figure}[htbp]
        \centering
        \captionsetup[subfigure]{justification=centering}
        \begin{subfigure}{0.45\textwidth}
          \includegraphics[width=\linewidth]{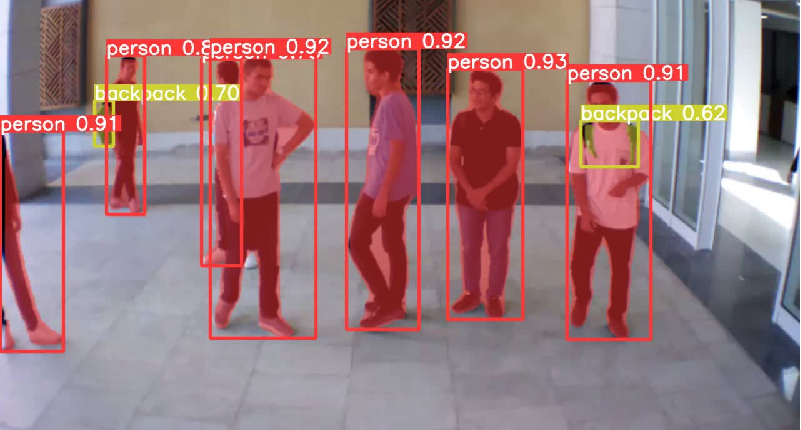}\par
          \caption{Yolov8 Segmentation.}
          \label{fig:seg_yolov8}
        \end{subfigure}
        \begin{subfigure}{0.45\textwidth}
          \includegraphics[width=\linewidth]{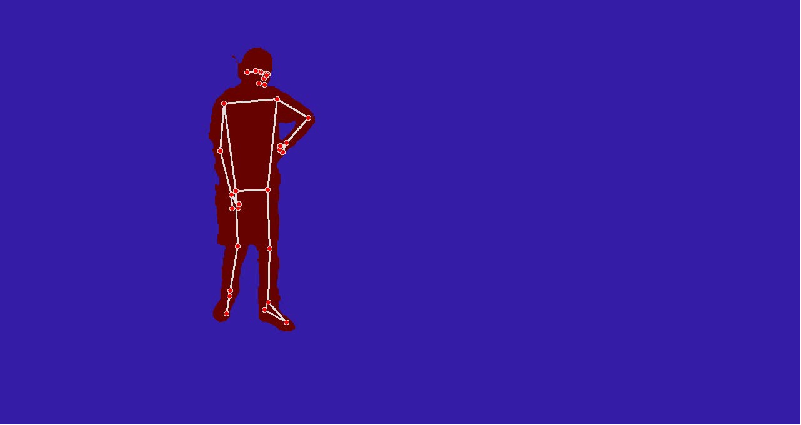} \par 
          \caption{Mediapipe Segmentation.}
          \label{fig:seg_mediapipe}
        \end{subfigure}
    \label{fig:Instance segmentation output}
    \caption{ \label{fig:Instance segmentation output} Instance segmentation output.}
    \end{figure}

    \subsubsection{YOLOv8 Segmentation}
    \par
        Within YOLOv8, a dedicated segmentation module is available, trained on the COCO dataset~\cite{lin2014microsoft}. This segmentation module demonstrates a noteworthy mean average precision (mAP) of 44.6. Moreover, it operates at an impressive processing speed of 67 frames per second. The results of YOLOv8 segmentation are visually presented in Fig.~\ref{fig:seg_yolov8}, showcasing the module's proficiency in accurately segmenting all subjects within the image.

    \subsubsection{MediaPipe Segmentation}
        MediaPipe, in addition to its primary function of pose estimation~\ref{sec: Pose Estimation}, offers the capability of subject segmentation. By enabling the \textit{enable\_segmentation} parameter within the pose model~\cite{mediapipe}, MediaPipe can detect and segment a single subject while predicting their pose, as shown in Fig.~\ref{fig:seg_mediapipe}. However, it faces limitations when detecting and segmenting multiple subjects within the same image, as observed in pose estimation. MediaPipe's design, tailored for single individual detection, presents challenges in simultaneous detection and segmentation of multiple subjects. In terms of accuracy and processing speed, MediaPipe achieves a segmentation accuracy rate of 96.21\% and processes at 21 frames per second. These results were obtained using their proprietary 'Selfie Dataset,' which is not publicly available~\cite{mediapipe}.
        
    \subsection{Face Detection}
    \label{sec: Face Detection}
        
        \par
        Face detection focuses on the identification and localization of human faces within images or video frames. By utilizing advanced algorithms and machine learning models, face detection can analyze pixel patterns in an image to detect facial features like eyes, nose, and mouth, enabling the determination of the presence of a human face. The importance of face detection becomes particularly evident in various applications, especially in the context of drones, particularly in surveillance scenarios~\cite{ding2016comprehensive}. 
        
        Within DroneVis, we provide diverse face detection models, each tailored to specific needs. Some of these models excel in identifying faces close to the camera, such as the MediaPipe Face model. Meanwhile, other models exhibit reliable performance in scenarios featuring faces at varying distances from the camera. This group includes the Haar Face Detector, Hog Face Detector, Dlib CNN Face Detector, and OpenCV DNN Face Detector. Furthermore, we offer a robust face detection model designed to handle distant faces, making it particularly suitable for drone applications where subjects are typically situated at a considerable distance from the drone. This model is YOLOv8 fine-tuned for face detection, ensuring high accuracy even in challenging scenarios.
        The outputs of the diverse face detection models are depicted in Fig.~\ref{fig:Output of various face detection models for near and distant faces}. In the first row, the results pertain to faces in close proximity to the camera, an instance where all models exhibit robust performance. Meanwhile, the second row showcases the outcomes for faces positioned at a considerable distance from the camera, featuring the outputs of the Haar Face Detector Model and the fine-tuned YOLOv8 Model. Notably, the remaining models encountered limitations in detecting faces at such distances and were consequently omitted from the figure.
        The subsequent sections provide a comprehensive exploration of the aforementioned models' characteristics and capabilities.

        \begin{figure*}
        \centering
        \captionsetup[subfigure]{justification=centering}
        \begin{subfigure}{0.45\textwidth}
          \includegraphics[width=\linewidth]{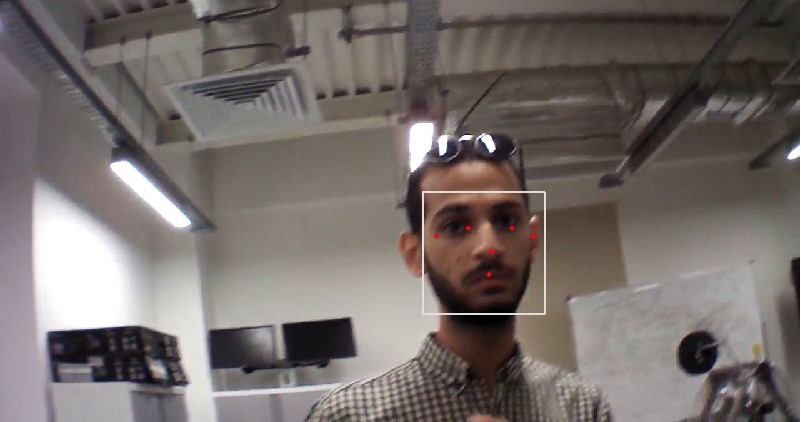}\par
          \caption{MediaPipe face model output for close faces.}
          \label{fig:MediaPipe Face model output for close faces}
        \end{subfigure}
        \begin{subfigure}{0.45\textwidth}
          \includegraphics[width=\linewidth]{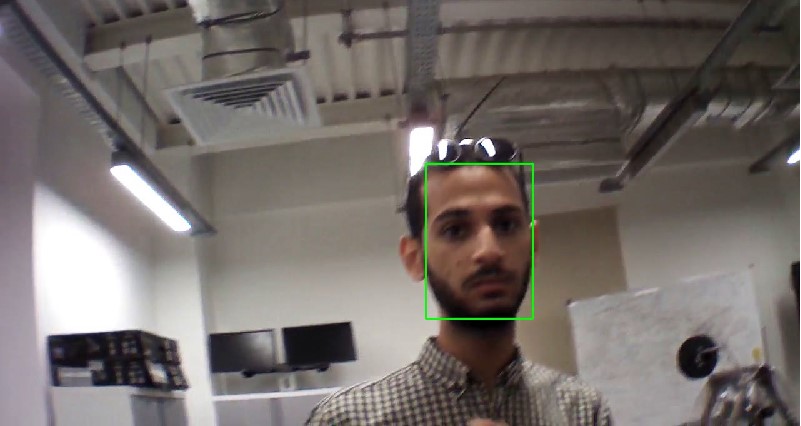} \par 
          \caption{Other models output for close faces.}
          \label{fig:Other models output for close faces}
        \end{subfigure}
        \begin{subfigure}{0.45\textwidth}
          \includegraphics[width=\linewidth]{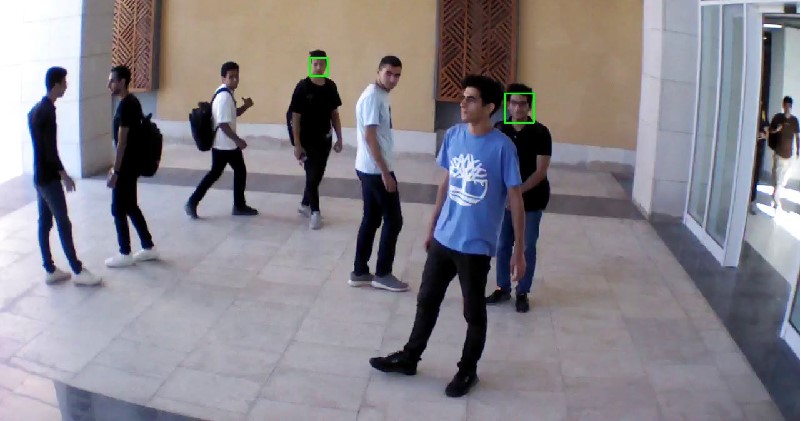}\par 
          \caption{Haar face detector output for far faces.}
          \label{fig:Haar Face Detector output for far faces}
        \end{subfigure}
        \begin{subfigure}{0.45\textwidth}
          \includegraphics[width=\linewidth]{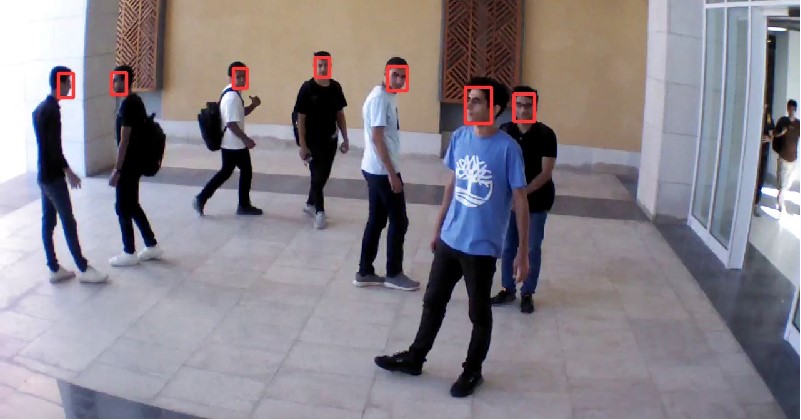}\par  
          \caption{Fine-tuned YOLOv8 for far faces.}
          \label{fig:Face Model pretrained on YOLOv8 for far faces}
        \end{subfigure}
        \label{fig:Output of various face detection models for near and distant faces}
        \caption{\label{fig:Output of various face detection models for near and distant faces} Output of various face detection models for near and distant faces.}
    \end{figure*}
         
        \subsubsection{MediaPipe Face}

        \par 

        MediaPipe Face Classifier, an advanced deep learning-based face detection approach by Google researchers~\cite{bazarevsky2019blazeface}. Using a multi-task learning framework, it optimizes face detection and classification tasks, combining CNNs and RNNs for feature extraction and face categorization. The method excels in diverse poses and lighting conditions, achieving 99.63\% accuracy with a low false positive rate of 0.17 on the "Labeled Faces in the Wild" (LFW) dataset~\cite{huang2007labeled}. Computationally efficient, suitable for deployment on resource-constrained devices like smartphones or drones, the method primarily focuses on close-up faces and may struggle with distant faces, limiting its applicability in scenarios such as drone-based applications
        
        \subsubsection{Haar Face Detector}

        \par 
        The Haar classifier is based on the Haar-like features proposed by Viola and Jones \cite{viola2001rapid}. These features are simple rectangular filters that capture specific patterns in an image, such as edges, lines, and corners. The classifier works by applying a series of these Haar-like features to sub-regions of an image and evaluating the response at each stage to determine the likelihood of a face being present. The Haar classifier consists of a cascade of weak classifiers, which are trained using a variant of the AdaBoost algorithm. Evaluated on ``Labeled Faces in the Wild'' (LFW) dataset \cite{huang2007labeled}, the classifier has demonstrated impressive results, achieving accuracies of over 95\% with low false-positive rates.

        \subsubsection{HOG Face Detector}

        \par 
        The Histogram of Oriented Gradients (HOG) \cite{dalal2005histograms} classifier is a popular face detection algorithm in computer vision. It captures gradient orientation distributions in image cells to represent features and uses a sliding window approach for detection. Compared to Haar classifiers, HOG is more robust to lighting and pose variations, detects faces of varying sizes, and offers computational efficiency for real-time applications. It achieved an accuracy of 95.6\% with a false positive rate of 0.13 on the "Labeled Faces in the Wild" (LFW) dataset. However, HOG requires ample training data and careful parameter selection. Despite its limitations, HOG remains widely used and effective in computer vision.

        \subsubsection{Dlib CNN Face Detector}

        \par 
        The Convolutional Neural Network (CNN) Face Detector in Dlib is a deep learning-based approach for detecting faces in images. Unlike traditional computer vision techniques that rely on hand-crafted features, CNNs learn to extract relevant features from the input data, which makes them more effective in dealing with complex tasks like face detection. It achieved an accuracy of 97.6\% on the LFW dataset. Additionally, the model is robust to variations in lighting, pose, and expression, making it suitable for real-world applications.

        \subsubsection{OpenCV DNN Face Detector}

        \par 
        The OpenCV library includes a deep neural network (DNN) based face detector, introduced in version 3.3. This detector utilizes the ResNet-10 architecture as its backbone and employs a single-shot multibox detector approach. It combines feature extraction using Haar-like features with classification using a CNN to identify faces in images. The model achieved an impressive accuracy of 99.60\% on the LFW dataset. However, one potential drawback is its higher computational cost compared to the Haar cascade classifier, making it less suitable for resource-constrained devices like smartphones or embedded systems.

       \subsubsection{Fine-tuned YOLOv8 for Face Detection}
        \par
        The previously mentioned face detection models demonstrated excellent performance for close-range faces. However, their effectiveness notably declined when confronted with distant faces. In response to this challenge, we have introduced an alternative face detection model, a fine-tuned version of YOLOv8 specifically optimized for face detection. We harnessed the renowned capabilities in object detection of YOLOv8 by tailoring the model to face detection through fine-tuning, where every layer of the YOLOv8 (nano version) model was made trainable. This process utilized a dedicated face detection dataset~\cite{fares_elmenshawii_face_detection_dataset}, which includes distant faces, making it especially more suitable for drone applications than the LFW dataset, which contains near faces only. The dataset encompasses approximately 16.7k images, with 80\% allocated for training and 20\% for testing. The model underwent training for 28 epochs, employing a meticulously chosen learning rate of 0.000713, determined through rigorous experimentation to optimize performance. The test results highlight the model's performance, achieving a precision of 0.89522, a recall of 0.80549, a mean Average Precision (mAP) of 0.88056, and an impressive operating speed of 55 frames per second (fps).
        
        In Fig.~\ref{fig:Output of various face detection models for near and distant faces}, it is evident that our fine-tuned YOLOv8 model excelled in detecting all faces positioned at a considerable distance from the camera. In contrast, the Haar face detector model managed to detect only two out of seven faces in the figure, while the other face detection models struggled to detect faces at such a significant distance from the camera.

    \subsection{Crowd Counting}
    \label{sec: Crowd Counting}   
    
        \par
        Crowd counting involves estimating the number of people in crowded scenes like urban areas, stadiums, and public events. Crowd counting is essential with drones for monitoring gatherings, managing crowd flow, and assessing affected populations for disaster response in case of emergency.
        
        \par
        Various approaches exist in crowd counting: detection-based, regression-based, and density-based regression counting. Detection-based methods involve people detection before counting, facing challenges in crowded scenes. Regression-based methods directly estimate the count but can be sensitive to variations and clutter. Density-based methods analyze overall crowd density, offering simplicity and efficiency but might lose accuracy in complex scenes. For low error rates and real-time performance, we adopt a cascaded convolutional neural network~\cite{sindagi2017cnn}. This leverages multi-task learning for crowd density map estimation and crowd count classification into distinct groups. Crowd count classification captures scene-level information, while density estimation focuses on local density, enhancing accuracy. Fig.~\ref{fig:Crowd counting output} illustrates an example of the model output.

    \begin{figure}[htbp]
        \centering
        \captionsetup[subfigure]{justification=centering}
        \begin{subfigure}{0.4\textwidth}
          \includegraphics[width=\linewidth]{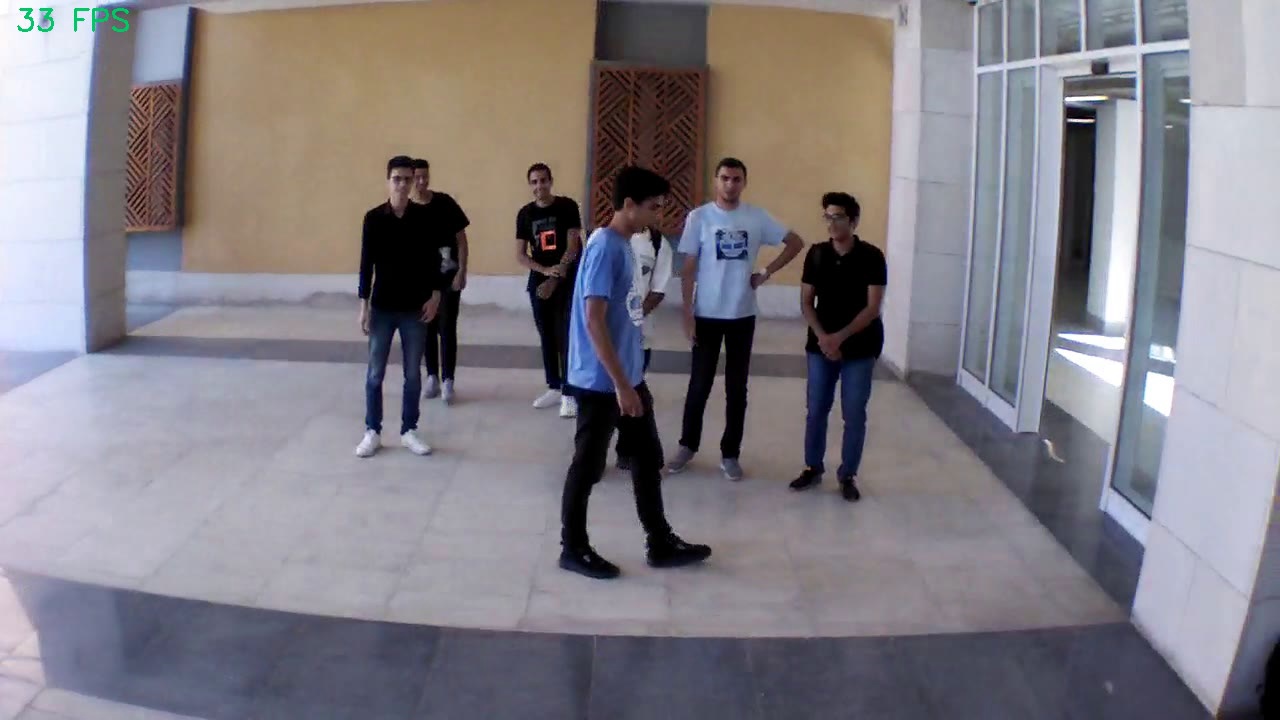}\par
          \caption{Input.}
          \label{fig:crowd_input}
        \end{subfigure}
        \begin{subfigure}{0.4\textwidth}
          \includegraphics[width=\linewidth]{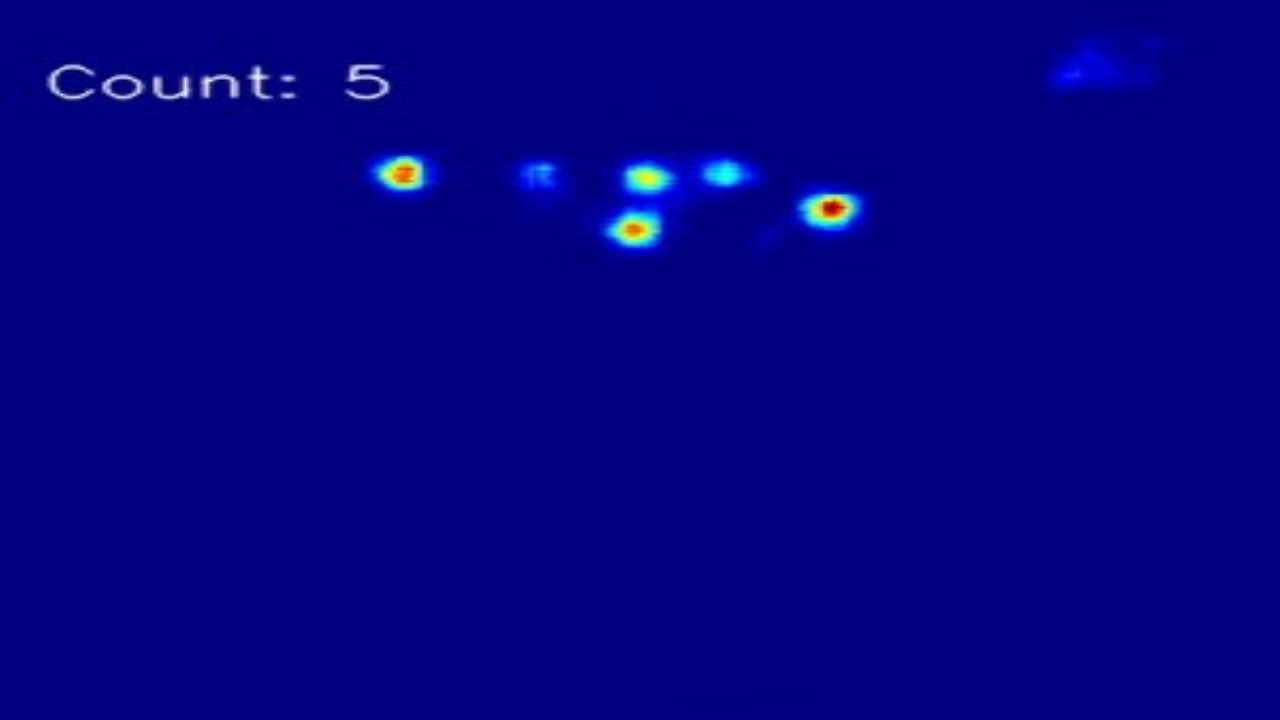} \par 
          \caption{Output.}
          \label{fig:mediapipe}
        \end{subfigure}
    \label{fig:Crowd counting output}
    \caption{\label{fig:Crowd counting output} Input and Output of Crowd counting.}
    \end{figure}

    \subsection{Action Recognition}
    \label {sec: Action Recognition} 
    
        \par
\begin{figure}[htbp]
        \centering
        \includegraphics[width=0.4\textwidth]{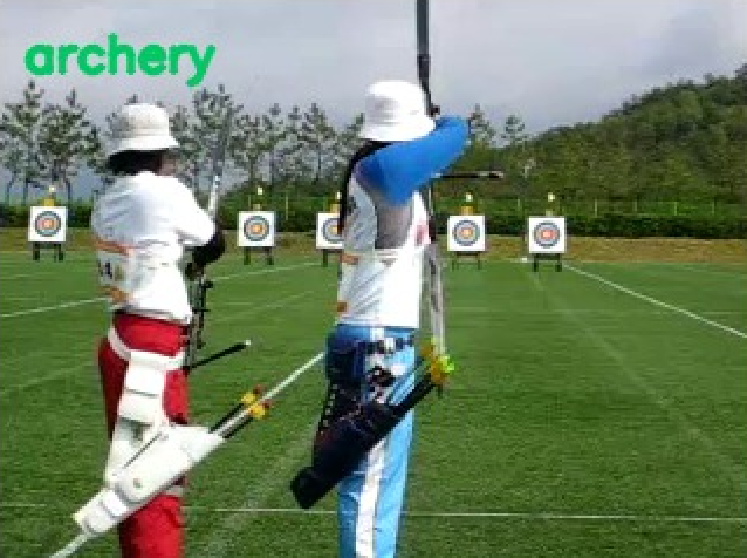}
        \caption{Action Recognition output. Source: UCF101 dataset~\cite{soomro2012ucf101}.}
        \label{fig:Action Recognition output}
        \end{figure}

        Action recognition involves the analysis of video sequences to identify and categorize specific human actions or activities. Action recognition with drones holds significant importance in various fields due to its potential to enhance safety and decision-making. By accurately identifying and tracking human actions or movements from an aerial perspective, drones can play a crucial role in security and surveillance operations, enabling rapid response to potential threats or emergencies. Additionally, drones equipped with action recognition capabilities in sports and entertainment can capture dynamic and engaging footage, enhancing the viewer experience. 
        For this task, we offer three different video recognition models that achieve state-of-the-art results on well-known video recognition datasets.  An example of the output of any of the action recognition models is shown in Fig.~\ref{fig:Action Recognition output}.

        \subsubsection{Time Space Transformer (TimeSformer)}

        \par 
        Instead of applying joint space-time attention, which is time-consuming, a divided space and time attention is proposed in ~\cite{bertasius2021space}. In this approach, each patch in an image is first used to compute temporal attention, with all patches having the same spatial index. The resulting encoding is then used for computing spatial attention with patches having the same temporal index. This approach surpasses the usage of spatial attention only or joint space-time attention in terms of accuracy. In addition, the inference cost of this approach is minimal compared to other well-known approaches relying on 3D convolution, such as SlowFast~\cite{feichtenhofer2019slowfast}.
        
        \subsubsection{Video Vision Transformer (ViViT)}

        \par 
        Four different architectures of video transformers are proposed in~\cite{arnab2021vivit} based on four different factorization techniques of spatiotemporal features. In the first variation (Spatiotemporal attention), the transformer encoder accepts different spatiotemporal portions of the video, performing joint space-time attention without factorization. In the second variation (Factorized encoder), a spatial transformer is applied first for all frames in a video, and then a temporal transformer deals with the learned encoding of each spatial transformer. In the third variation (Factorized self-attention), a similar approach to the first variation is applied but with factorization into spatial attention and temporal attention. In the final variation (Factorized dot-product attention), the factorization is applied to the multi-head dot-product attention operation. The first variation has the highest inference cost and achieves the best results on the Kinetics-400 dataset~\cite{kay2017kinetics}. On the other hand, the second variation has the minimal inference cost compared to the other variations, with a slight decrease in accuracy compared to the first variation.
        
        \subsubsection{Video Masked Autoencoders (VideoMAE)}

        \par 
        A tube masking strategy is proposed in~\cite{tong2022videomae} to force the network to learn important features in a video, which is achieved by masking a large portion of cubes in the video - cube embedding is used to represent one token - with negligible motion, reducing the computation cost and enhancing the model performance. The few remaining unmasked tokens are used with the vanilla Vision Transformer (ViT) to capture the spatiotemporal relation between them. This approach achieved state-of-the-art results on the Kinetics-400 dataset~\cite{kay2017kinetics} with 87.4\% accuracy, compared to 80.7\% and 84.9\% accuracy of TimeSformer and ViViT, respectively. In terms of inference cost, VideoMAE requires 7397 GFLOPS, while TimeSformer and ViViT require 8353 and 3981 GFLOPS, respectively~\cite{tong2022videomae}.

    \subsection{Depth Estimation}
    \label{sec: Depth Estimation}

        \par 
        Monocular depth estimation means deducing the distance or depth information of objects within a scene from a single 2D image or video frame. This process facilitates comprehension of the spatial arrangement of the environment and the relative separations between objects. Crucial for drones in autonomous navigation, obstacle avoidance, and scene understanding, monocular depth estimation extracts cues from the drone's camera feed; the drone gains the ability to perceive its surroundings, gauge the elevation of obstacles or landmarks, and make informed decisions to ensure secure and efficient flight. 

\begin{figure}[htbp]
          \centering
          \includegraphics[width=0.5\textwidth]{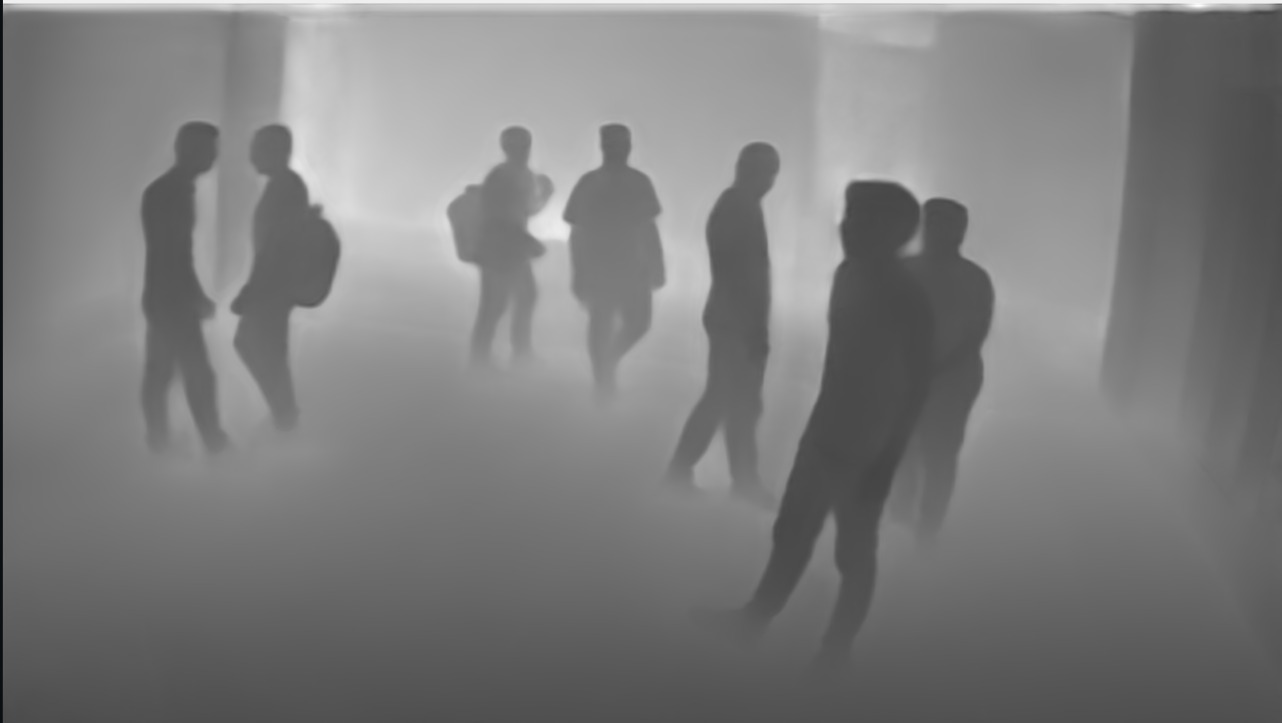}
          \caption{Depth estimation output.}
          \label{fig:Depth estimation output}
        \end{figure}

        The method in~\cite{kim2022global} is employed, featuring an encoder with transformers for a large receptive field and a decoder with skip connections and selective feature fusion using attention maps. This preserves structural details and generates a detailed feature map. The approach was evaluated on KITTI dataset~\cite{geiger2013vision}, achieving 0.908 $\delta_{1}$, representing the percentage of pixels with a relative depth error below a threshold of 1.25. Fig.~\ref{fig:Depth estimation output} illustrates an example output, where darker regions denote nearer objects and brighter regions represent farther objects.
        
    \subsection{Road Segmentation and Lane Detection}
    \label{sec: Road segmentation and Lane Detection}

        \par 
        Drones equipped with road segmentation and lane detection capabilities significantly enhance their surrounding awareness, enabling their application in diverse scenarios such as search and rescue operations, traffic management, and surveillance tasks. In this context, we employed YOLOP~\cite{wu2022yolop}, a panoptic driving perception network designed to simultaneously address multiple tasks, including object detection, deliverable area segmentation, and lane detection.

        \par 
        The architecture of YOLOP comprises a shared encoder, leveraging the CSPDarknet backbone~\cite{wang2021scaled}, which is augmented by Spatial Pyramid Pooling (SSP)~\cite{he2015spatial} and Feature Pyramid Network~\cite{lin2017feature} modules. These additions facilitate the generation of multi-scale features with varying semantic levels. The encoder's output is then directed to a decoder, which includes three separate heads dedicated to each of the three tasks. The model is trained end-to-end for the three tasks simultaneously.

        \par
\begin{figure}[htbp]
        \centering
        \captionsetup[subfigure]{justification=centering}
        \begin{subfigure}{0.4\textwidth}
          \includegraphics[width=\linewidth]{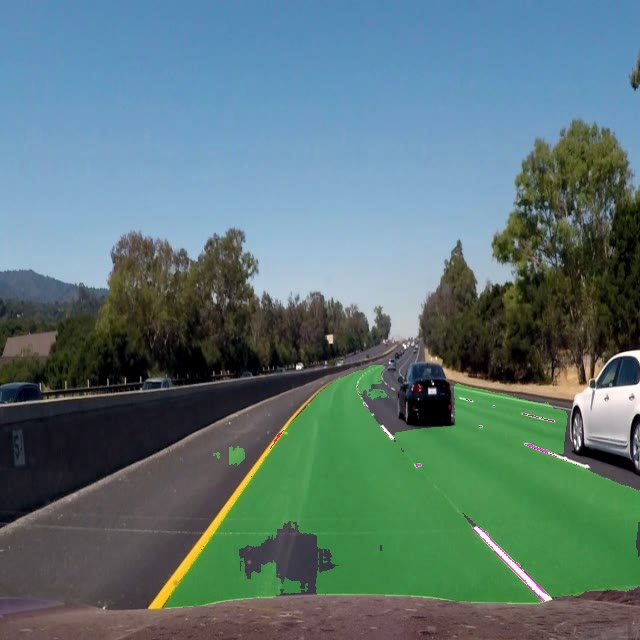}\par
          \caption{Driveable area segmentation output.}
          \label{fig:road}
        \end{subfigure}
        \begin{subfigure}{0.4\textwidth}
          \includegraphics[width=\linewidth]{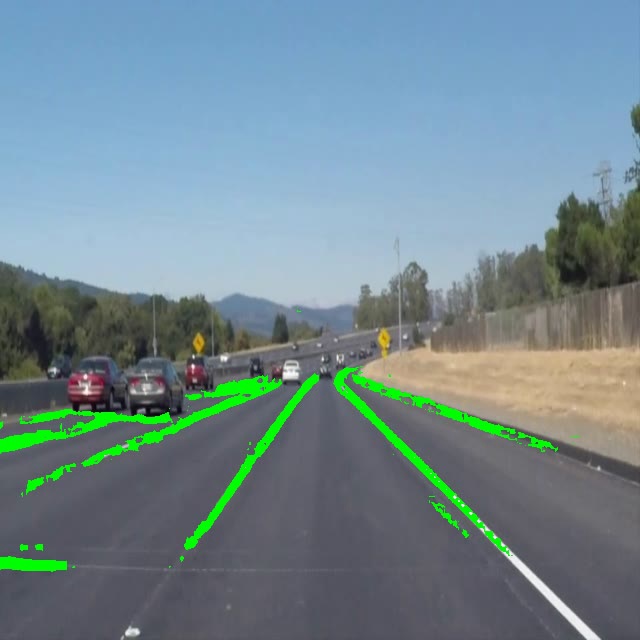} \par 
          \caption{Lane detection output}
          \label{fig:lane}
        \end{subfigure}
        \label{fig:Output of road and lane detection}
        \caption{Output of road segmentation and lane detection. Source: Kaggle KITTI Road Segmentation Dataset available at \url{https://www.kaggle.com/datasets/sakshaymahna/kittiroadsegmentation.}}
        \end{figure}

 
        The performance evaluation of the YOLOP model was conducted using the BDD100k dataset~\cite{yu2018bdd100k}, yielding commendable results across the board. Notably, the model achieved a road segmentation accuracy of 91.5\% mean Intersection over Union (mIOU) and a lane detection accuracy of 70.50\%, with a remarkably real-time processing speed of 17 frames per second (fps). Figures~\ref{fig:road}, and~\ref{fig:lane} show the output of the driveable area segmentation head and the lane detection head simultaneously.

    \subsection{Our Recommendation}
    \label{sec:Our Recommendation}

        \par 
        This section presents a comparative analysis of previously discussed models, offering a recommended model for each task based on a balance between model performance and speed. Initially, these recommended models are pre-configured as the default choices within the DroneVis library. However, users retain the flexibility to tailor these defaults according to their individual preferences.

        \par 
        Table~\ref{tab: Performance Comparison of Various Models} summarizes the aforementioned tasks and associated models. The table provides insights into their individual performance metrics, frames per second (FPS), the platform they are compatible with, and the size of their pre-trained weights. These evaluations were conducted on a laptop equipped with an Intel Core 8 CPU. The dfault models for each task in DroneVis are highlighted in bold.
        In this context, our attention is directed toward tasks that offer multiple model options. We establish the criteria guiding our recommendations for these specific tasks:
        \begin{itemize}
            \item For object detection, we advocate the use of the YOLOv8 model, as it demonstrates exceptional speed and accuracy, achieving 80 FPS and 53.9 AP.
            
            \item In the domain of pose estimation and instance segmentation, YOLOv8 is also our top recommendation, primarily due to its superior speed compared to MediaPipe. Additionally, YOLOv8 performs effectively when dealing with multiple individuals, in contrast to MediaPipe, which is limited to providing pose estimations and segmentation for a single subject.
            
            \item Regarding action recognition, we suggest the adoption of VideoMAE. This choice is based on the model's more compact weight size when compared to other action recognition models. Moreover, VideoMAE stands out with its exceptional accuracy while maintaining a competitive processing speed. 

            \item For face detection, we suggest the adoption of the fine-tuned YOLOv8 model because of its reliable performance with distant faces compared to other face detection models. In addition, it has the highest processing speed.
        \end{itemize}


        \begin{table}[ht]
        \centering
        \caption{Performance Comparison of Various Models in the Computer Vision Tasks Implemented in Drone-Vis. Default models are in bold font.}
        \label{tab: Performance Comparison of Various Models}
        \resizebox{\linewidth}{!}{
        \tabularnewline
        \begin{tabular}{|c|c|c|c|c|c|c|c|}
        \hline
        \textbf{Task} & \textbf{Models} & \textbf{Framework} & \textbf{FPS} & \textbf{Evaluation Score} & \textbf{Weights Size} & \textbf{Quantized} & \textbf{Dataset} \\ \hline
        Object Detection & Faster R-CNN & Pytorch & 15 & 42.7 AP& 77 MB & & COCO \\ \cline{2-8}
        & SSD & Pytorch & 24 &   46.5 AP & 14 MB & & COCO \\ \cline{2-8}
        & YOLOv5 & Pytorch & 20 & 50.7 AP & 186MB (xlarge) & & COCO\\ \cline{2-8}
        & \textbf{YOLOv8} & Pytorch & 80 & 53.9 AP & 50MB (medium) & \textcolor{green}{\ding{52}} & COCO\\ \hline
        \hline
        Object Tracking & \textbf{YOLOv8} & Pytorch & 53 & 77.8 (MOTA) & 24MB (small) & \textcolor{green}{\ding{52}} & COCO \\ \hline
        \hline
        Instance Segmentation & \textbf{YOLOv8} & Pytorch & 46 & 44.6 AP & 24MB (small) & \textcolor{green}{\ding{52}} & COCO \\ \cline{2-8}
        & MediaPipe & TF Lite & 21 & 96.21\% & 249 KB &\textcolor{green}{\ding{52}} & Selfie Dataset \\ 
        
        \hline
        \hline
        Pose Estimation & \textbf{YOLOv8} & Pytorch & 67 & 50.4 AP & 7MB (nano)& \textcolor{green}{\ding{52}} & MPII \\ \cline{2-8}
        & MediaPipe & TF Lite & 30 & 53.8 mAP & 6MB (lite)& \textcolor{green}{\ding{52}} & MPII \\ \hline
        \hline
        Face Detection & MediaPipe & TF Lite & 200 & 99.6\% & 225KB &\textcolor{green}{\ding{52}} & LFW \\ \cline{2-8}
        & Haar & OpenCV & 15 & 95\% & 900KB & & LFW \\ \cline{2-8}
        & HOG & Dlib & 35 & 95.5\% & 127KB & & LFW \\ \cline{2-8}
        & DLib CNN & Dlib & 4 & 97.6\% & 713KB & & LFW \\ \cline{2-8}
        & OpenCV DNN & OpenCV & 45 & 99.6\% & 5.4MB & & LFW \\ \cline{2-8}
        & \textbf{Fine-tuned YOLOv8} & Pytorch & 55 & 85.79\% & 36.6 MB & \textcolor{green}{\ding{52}} & Face Detection Dataset\\ \cline{2-8}
        \hline
        \hline
        Crowd Counting & \textbf{Cascaded MTL} & Pytorch & 11 & 101 MAE & 10MB & & UCY \\ \hline
        \hline
        Action Recognition (16 frames) & TimeSformer & Pytorch & 4 & 77.9\% & 487MB & & ActivityNet \\ \cline{2-8}
        & ViViT & Pytorch & 5 & 84.9\% & 356MB & & YouTube-VOS \\ \cline{2-8}
        & \textbf{VideoMAE} & Pytorch & 4 & 87.4\% & 346MB & & UCF101 \\ \hline
        \hline
        Depth Estimation & \textbf{GLPDepth} & Pytorch & 4 & 0.908 $\delta_{1}$ & 245MB & & KITTI \\ \hline
        Road Segmentation & \textbf{YOLOP} & Pytorch & 17 & 75 mAP & 36MB (onnx) & & BDD100k \\ \hline
        \end{tabular}
        }
        \end{table}

\section{Comparison to Related Software}
\label {sec:Comparison to Related Software}

    \par 
    There is limited prior research related to the application of computer vision algorithms on Parrot drones. The work in~\cite{rohan2019convolutional} focuses on implementing person detection and tracking on Parrot drones, achieving a frame rate of 58 frames per second. Similarly, the Deep Drone project by Han et al.~\cite{han2016deep} also addresses these tasks, employing Faster R-CNN for detection and the KFC algorithm for tracking at frame rates of 1.6 and 70 frames per second, respectively. However, it's worth noting that neither of these previous works provides a library for their research, and their source code is not publicly accessible~\cite{rohan2019convolutional, han2016deep}. This lack of accessibility restricts the ability of other researchers and developers to contribute to or build upon their findings.

    \par 
    In contrast to these previous studies, the MMCV library~\cite{MMCV_Contributors_OpenMMLab_Computer_Vision_2018} serves as a valuable point of reference for comparison. MMCV is designed for a wide range of computer vision tasks, including image and video processing, data visualization, various convolutional neural network architectures, and customized activation functions. Additionally, MMCV offers support for multiple tasks such as classification, detection, segmentation, pose estimation, and action recognition. It is implemented on the PyTorch framework and is compatible with various operating systems, including Windows, Linux, and MacOS. The main drawbacks of the MMCV library are that it lacks the presence of user interfaces such as CLI and GUI and requires a good knowledge of the previously mentioned tasks to select the suitable model among the variety of models they provide for each task. These drawbacks are not found in the DroneVis library, as we offer multiple user interfaces to make the library user-friendly and suitable for any user, regardless of their programming knowledge, as depicted in figures~\ref{fig:Drone-Vis GUI} and~\ref{fig:Drone-Vis CLI}.
    Furthermore, we provide a default model for each task in DroneVis, suitable for users who may lack familiarity with the evaluation metrics of various computer vision models and might be unsure about model selection criteria. Additionally, we offer users the flexibility to change the default model based on their preferences. 
    
    Table~\ref{tab:Comparison to related software} presents a comprehensive comparison between DroneVis and MMCV libraries, encompassing the aforementioned aspects and delving into key factors. These factors include:
    \begin{itemize}
        \item Test Coverage: Test coverage is a metric that measures the proportion of a codebase covered by automated tests, enhancing code reliability by identifying untested code paths. Both libraries demonstrate commendable test coverage, assuring code quality.

        \item Type Checking: Type checking is implemented in both libraries to maintain code consistency and prevent errors.

        \item Issue and Pull Request (PR) Templates: These templates streamline contributions, simplifying the process for users to report issues or submit PRs.

        \item Recent Activity (Measured by Last Commit): Active development is evident, with both libraries having a last commit as recent as one day ago.

        \item Approved PRs (Signifying Active Community Engagement): The number of approved Pull Requests in the past six months showcases active community engagement, with DroneVis having 36 and MMCV 64 approved PRs.
    \end{itemize}
    These factors collectively contribute to the libraries' reliability, accessibility, and vitality of their respective communities.

    \begin{table}[t!]

    \centering
    \caption{Comparison to related software}
    \label{tab:Comparison to related software}
    \begin{tabular}{lcc}
    \toprule
    & Drone-Vis  & MMCV \\
    \midrule
    Backend & TFLite/Pytorch &  Pytorch\\
    
    User Guide/Tutorials & \textcolor{green}{\ding{52}}/ \textcolor{green}{\ding{52}} & \textcolor{green}{\ding{52}}/ \textcolor{green}{\ding{52}}\\
    
    API Documentation & \textcolor{green}{\ding{52}} & \textcolor{green}{\ding{52}}\\
    
    Pretrained Models & \textcolor{green}{\ding{52}} & \textcolor{green}{\ding{52}}\\
    Default Models  & \textcolor{green}{\ding{52}} & \textcolor{red}{\ding{56}} \\
    CLI/GUI & \textcolor{green}{\ding{52}} / \textcolor{green}{\ding{52}} & \textcolor{red}{\ding{56}}/\textcolor{red}{\ding{56}} \\
    Test Coverage & 83\% & 69\% \\
    Type Checking & \textcolor{green}{\ding{52}} & \textcolor{green}{\ding{52}}\\
    Issue PR Template & \textcolor{green}{\ding{52}} & \textcolor{green}{\ding{52}} \\
    Last Commit (age) & 1 day & 1 day\\
    Approved PRs (6 month) & 40 & 64\\
    \bottomrule
    \end{tabular}
    \end{table}

\section{Vision and Drone Integration}
\label{sec:Vision and Drone Integration}

    DroneVis excels in its seamless integration of computer vision models with drone technology, enabling intelligent and autonomous scene tracking. An illustrative example is its integration of the YOLOv8 object detection model with drones for autonomous subject tracking. This approach involves real-time analysis of video frames captured by the drone's camera. YOLOv8 then predicts bounding box coordinates for detected objects, enabling the drone to take appropriate actions based on object position and size. This integration not only underscores DroneVis's adaptability but also opens the door to incorporating various vision models, offering endless potential for precise and automated computer vision tasks in real-world applications.
    
\begin{figure}[htbp]
    \centering
    \captionsetup[subfigure]{justification=centering}
    \begin{subfigure}{0.45\textwidth}
      \includegraphics[width=\linewidth]{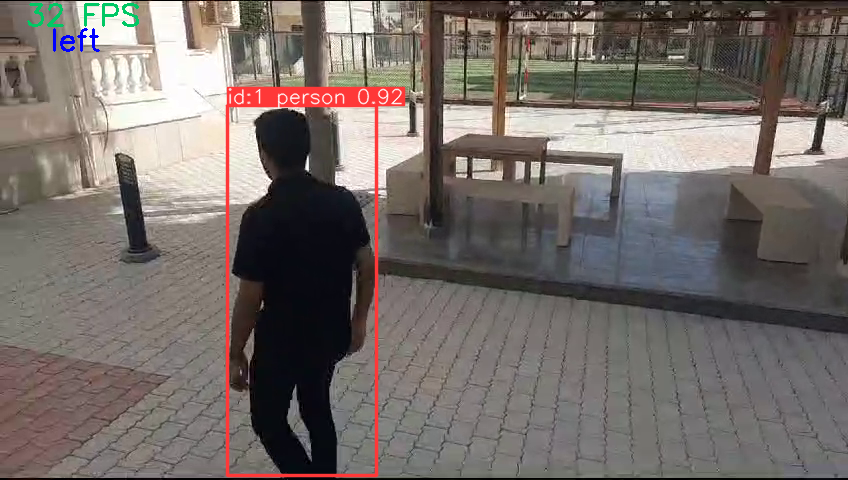}\par
      \caption{Drone steering left.}
      \label{fig:task-left}
    \end{subfigure}
    \begin{subfigure}{0.45\textwidth}
      \includegraphics[width=\linewidth]{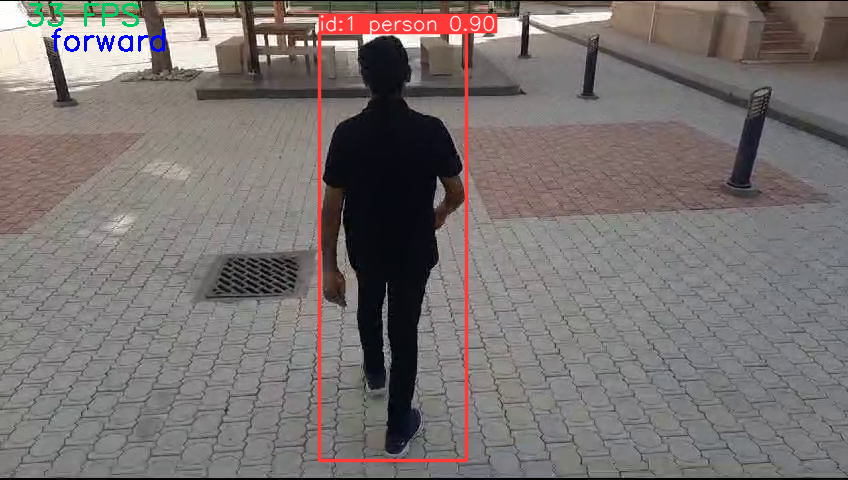} \par 
      \caption{Drone advancing.}
      \label{fig:task-forward}
    \end{subfigure}
    \caption{Output of the drone tacking for left and forward cases.}
    \label{fig:task-navigation}
    \end{figure}

   Here is a code snippet, accompanied by its output in Fig.~\ref{fig:task-navigation}, demonstrating just how effortlessly DroneVis facilitates the integration of drone control with computer vision models. The process is user-friendly, allowing developers to concentrate on crafting intelligent drone applications without being burdened by the technical complications of drone communication and control.

    \begin{mintedbox}
    import cv2
    from dronevis.drone_connect import Drone
    from dronevis.models import YOLOv8Detection
    from dronevis.utils.general import write_on_image
    
    drone = Drone(); drone.connect() # Connect to drone network
    model = YOLOv8Detection(track=True)

    EPS_HORIZONTAL = 0.1; EPS_VERTICAL = 0.3; EPS_HEIGHT = 0.3
    
    def take_action(results):
        boxes = results[0].boxes
        x, y, _, h = boxes[0].xywhn.detach().cpu().numpy()[0]
    
        if (abs(x - 0.5) <= EPS_HORIZONTAL and abs(y - 0.5) <= EPS_VERTICAL
            and abs(h - 0.5) <= EPS_HEIGHT):
            drone.hover(); return "hover"
    
        if x >= 0.5 + EPS_HORIZONTAL:
            drone.right(); return "right"
    
        if x <= 0.5 - EPS_HORIZONTAL:
            drone.left(); return "left"
    
        if y >= 0.5 + EPS_VERTICAL:
            drone.downward(); return "down"
    
        if y <= 0.5 - EPS_VERTICAL:
            drone.upward(); return "up"
    
        if h >= 1 + EPS_HEIGHT:
            drone.backward(); return "backward"
    
        drone.forward(); return "forward"
    
    def operating_callback(frame, _):
        results = model.raw_predict(frame, classes=0, verbose=False)
        action = "hover"
        if len(results[0].boxes) > 0: action = take_action(results)
        frame = write_on_image(results[0].plot(), action)
        cv2.imshow("frame", frame); cv2.waitKey(1)
    
    model.load_model()
    none_lambda = lambda: None # dummy callback for closing the video
    drone.connect_video(none_lambda, operating_callback, "None")
    \end{mintedbox}

    Furthermore, this example highlights the potential for extending DroneVis to more advanced tasks, such as Simultaneous Localization and Mapping (SLAM). By integrating additional components and algorithms into DroneVis, developers can create drones that not only track objects but also navigate complex environments, map their surroundings, and execute more advanced missions autonomously. This showcases the versatility and adaptability of DroneVis for various drone applications and underscores its role as a powerful tool for pushing the boundaries of drone technology.

\section{Conclusion and Future Work}
\label{sec:Conclusion and Future Work}

    \par 
    In this work, we have introduced the DroneVis library, which serves as a comprehensive solution for automating computer vision algorithms on Parrot drones. The library boasts a wide range of features, including comprehensive documentation, a demonstrative testing environment, adherence to coding standards, and a reliable test coverage percentage.
    Furthermore, DroneVis offers diverse user interfaces, encompassing both graphical and command-line interfaces. Notably, it also enables drone control through innovative methods such as hand gesture recognition, significantly expanding the range of drone applications.

    \par 
    The library supports an extensive selection of computer vision tasks, making Parrot drones suitable for applications ranging from surveillance and security to rescue and disaster management. Additionally, for each task, users have the flexibility to choose from a variety of neural network models, tailoring their approach to their specific preferences.

    \par 
    For future work, our primary objective is to integrate a localization module into the library. This enhancement will empower drones with autonomous navigation capabilities, opening up new horizons for their utilization. In addition an module for ensuring the automo

\bibliographystyle{ACM-Reference-Format}
\bibliography{sample-base}

\end{document}